%% file: main.tex
\documentclass[10pt,twocolumn,letterpaper,table]{article}

\pdfoutput=1

\usepackage[pagenumbers]{cvpr} 
\usepackage{pifont}

\input{preamble}

\definecolor{cvprblue}{rgb}{0.21,0.49,0.74}
\usepackage[pagebackref,breaklinks,colorlinks,allcolors=cvprblue]{hyperref}

\def\paperID{13617} 
\def\confName{CVPR}
\def\confYear{2025}
\def\ours{FastVLM}
\def\oursImgEnc{FastViTHD}
\newcommand{\cmark}{\ding{51}}%
\newcommand{\xmark}{\ding{55}}%

\title{\ours{}: Efficient Vision Encoding for Vision Language Models}

\author{Pavan Kumar Anasosalu Vasu$^\star$$^\dagger$
\and
Fartash Faghri$^\star$
\and
Chun-Liang Li$^\star$
\and
Cem Koc$^\star$
\and
Nate True
\and
Albert Antony
\and
Gokul Santhanam
\and
James Gabriel
\and
Peter Grasch
\and
Oncel Tuzel$^\star$
\and
Hadi Pouransari$^\star$$^\dagger$
\and
Apple \\
{\tt\small \{panasosaluvasu,fartash,chunliang\_li,cem\_koc,otuzel,mpouransari\}@apple.com} \\
\small{$^\star$Core authors; $^\dagger$Project lead}
}

\begin{document}
\maketitle
\input{sec/0_abstract}    
\input{sec/1_intro}

\input{sec/2_related_works}
\input{sec/3_method}

\input{sec/4_experiments}

\input{sec/5_conclusion.tex}
\input{sec/6_acknowledgement.tex}
{
    \small
    \bibliographystyle{ieeenat_fullname}
    \bibliography{main}
}

\input{sec/X_suppl}

\end{document}

%% file: preamble.tex
\usepackage{graphicx}
\usepackage{multirow}
\usepackage{amssymb}
\usepackage{lipsum}
\newcommand{\red}[1]{{\color{red}#1}}
\newcommand{\green}[1]{{\color{green}#1}}


%% file: sec/0_abstract.tex
\begin{abstract}

Scaling the input image resolution is essential for enhancing the performance of Vision Language Models (VLMs), particularly in text-rich image understanding tasks. However, popular visual encoders such as ViTs become inefficient at high resolutions due to the large number of tokens and high encoding latency. At different operational resolutions, the vision encoder of a VLM can be optimized along two axes: reducing encoding latency and minimizing the number of visual tokens passed to the LLM, thereby lowering overall latency. Based on a comprehensive efficiency analysis of the interplay between image resolution, vision latency, token count, and LLM size, we introduce FastVLM—a model that achieves an optimized trade-off between resolution, latency, and accuracy. FastVLM incorporates \oursImgEnc{}, a novel hybrid vision encoder designed to output fewer tokens and significantly reduce encoding time for high-resolution images. Unlike previous methods, FastVLM achieves the optimal balance between visual token count and image resolution solely by scaling the input image, eliminating the need for additional token pruning and simplifying the model design. In the LLaVA-1.5 setup, FastVLM achieves 3.2$\times$ improvement in time-to-first-token (TTFT) while maintaining similar performance on VLM benchmarks compared to prior works. Compared to LLaVa-OneVision at the highest resolution (1152$\times$1152), FastVLM achieves better performance on key benchmarks like SeedBench, MMMU and DocVQA, using the same 0.5B LLM, but with 85$\times$ faster TTFT and a vision encoder that is 3.4$\times$ smaller. Code and models are available at \url{https://github.com/apple/ml-fastvlm}.

\end{abstract}

%% file: sec/1_intro.tex
\section{Introduction}
\label{sec:intro}

\begin{figure}
\centering
\begin{subfigure}[b]{0.45\textwidth}
   \includegraphics[width=1\linewidth]{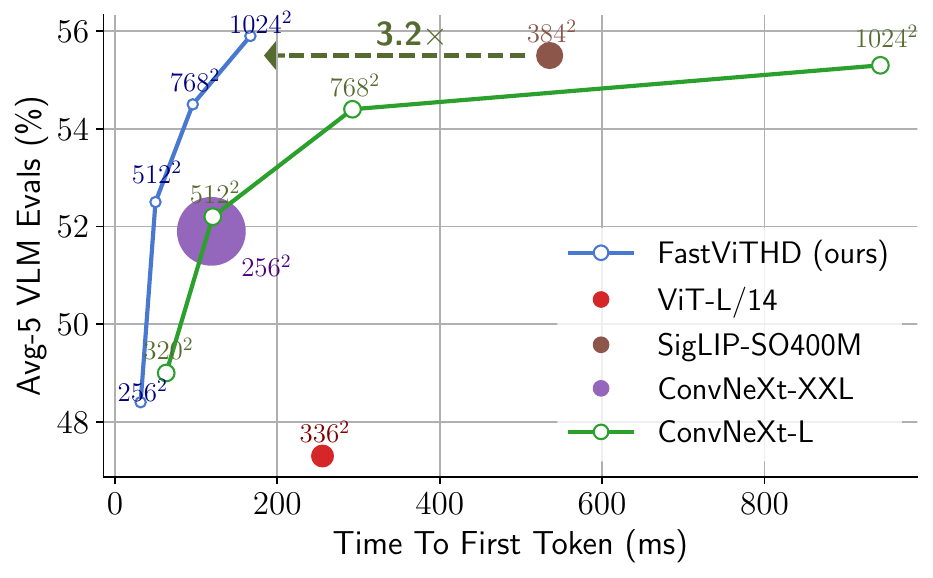}
\vspace*{-15pt}
   \caption{Qwen2-0.5B}
   \label{fig:acc_v_latency_qwen} 
\end{subfigure}
\vspace{10pt}

\begin{subfigure}[b]{0.45\textwidth}
   \includegraphics[width=1\linewidth]{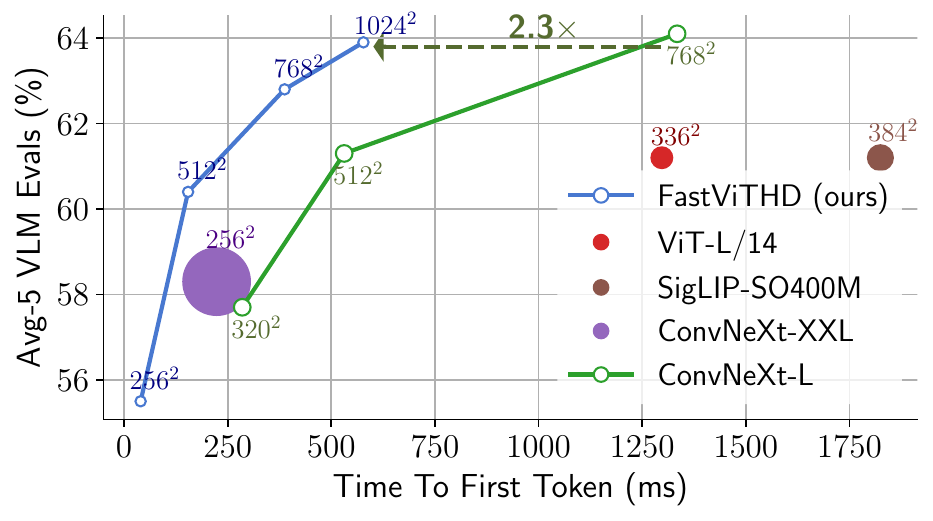}
\vspace*{-15pt}
   \caption{Vicuna-7B}
   \label{fig:acc_v_latency_vicuna}
\end{subfigure}
    \vspace{-5pt}

\caption{\textbf{FastVLM is more than 3$\times$ faster than prior work.} Comparison of commonly used vision encoders for VLMs with (a) Qwen2~\cite{qwen2} 0.5B LLM and (b) Vicuna 7B~\cite{zheng2023judging} LLM. All the vision encoders are CLIP~\citep{CLIP} pretrained. For a fair comparison all models are trained using LLaVA-1.5~\cite{liu2023improvedllava} setup with the vision encoders made trainable for resolution adaptation, see \cref{sec:experiments} for more details. Marker size for each model corresponds to number of parameters of the vision encoder. The $x$-axis is the sum of vision encoder latency and LLM prefilling time. All models are benchmarked on an M1 Macbook Pro. }
\label{fig:acc_vs_latency}
    \vspace*{-5pt}
\end{figure}

Vision Language Models (VLMs) enable visual understanding alongside textual inputs. VLMs are often built by passing visual tokens from a pretrained vision backbone to a pretrained Large Language Model (LLM) through a projection layer. Previous works~\citep{liu2023llava, liu2023improvedllava} have explored various training and fine-tuning strategies for these three components: the vision backbone, the projection layer, and the LLM, which is typically a decoder-only transformer~\citep{vaswani2017transformer} model.

Several studies~\citep{luo2024feast, ge2024convllava, mckinzie2024mm1methodsanalysis} highlight image resolution as a key factor in VLM performance, especially for text and chart-rich data. However, increasing image resolution presents multiple challenges. First, pretrained vision encoders may not support high-resolution images, as this would make pretraining inefficient. To address this, one approach is to continuously  pretrain the vision backbone to adapt it for high resolutions~\citep{beyer2024paligemmaversatile3bvlm}. Alternatively, tiling strategies, such as Sphinx~\citep{lin2023sphinx}, S2~\citep{shi2024we}, and AnyRes~\citep{liu2023improvedllava}, divide images into subregions, with each subregion processed independently by the backbone. 

A further challenge is the runtime computational cost associated with high-resolution inference. Both single high-resolution inference and multiple inferences at lower resolution (the tiling strategy) result in significant latency when generating visual tokens. Additionally, high-resolution images naturally produce more tokens, which increases the LLM prefilling time (the LLM forward pass time on all tokens in the context, including visual tokens), thereby further increasing the time-to-first-token (TTFT), which is the sum of the vision encoder latency and the LLM prefilling time.

In this work, we study VLM design and training from a runtime efficiency perspective.
We explore the optimization landscape as image resolution increases, aiming to improve accuracy-latency trade-off, where latency includes both the vision encoder inference time and the LLM prefilling time. Using extensive experiments with different LLM sizes and resolutions, we establish the Pareto optimal curve for a specific vision backbone, showing the best accuracy achievable within a given runtime budget (TTFT) based on different choices of resolution and LLM size.

We start by exploring the use of a hybrid convolutional-transformer architecture FastViT~\citep{vasu2023fastvit}, pretrained with MobileCLIP~\citep{mobileclip2024}, as a vision backbone for the VLM setup (\Cref{sec:fastvit}). We demonstrate the potential of this hybrid backbone, which generates visual tokens over 4$\times$ faster than a ViT model while achieving higher overall VLM accuracy with multi-scale features  (\Cref{sec:multi-scale}). 

However, further architectural optimization is possible when the primary goal is a high-resolution VLM (rather than embedding generation as in MobileCLIP-pretrained FastViT). We introduce a new hybrid vision encoder, \oursImgEnc{}, specifically designed for efficient VLM performance on high-resolution images (\Cref{sec:new-arch}), and use it as the vision backbone to obtain FastVLM through visual instruction tuning. 
FastVLM demonstrates a significantly improved accuracy-latency trade-off over VLMs based on ViTs, convolutional encoders, and our previously discussed hybrid FastViT for different input image resolutions and LLM sizes (\Cref{fig:acc_v_latency_qwen}, \cref{fig:acc_v_latency_vicuna,fig:dec_enc}).
In particular, FastVLM outperforms several prior works while being smaller, faster, and trained with less data (\Cref{tab:main_result}). Compared to LLaVA-OneVision~\citep{li2024llava} operating at the highest possible resolution (1152$\times$1152), FastVLM obtains comparable performance with the same 0.5B LLM, but with 85$\times$ faster TTFT and a 3.4$\times$ smaller vision encoder.

The following is a summary of our contributions:
\begin{itemize}
    \item We show that hybrid vision backbones outperform ViTs in VLMs, and introduce additional architectural interventions, such as multi-scale vision features, to further improve VLM performance while maintaining efficiency.
    \item We design and pretrain a new hybrid architecture, \oursImgEnc{}, optimized for efficient VLM performance with high resolution input for FastVLM. In a controlled experimental setup, where only the vision backbone is changed, we show that \oursImgEnc{} outperforms its ViT-based and convolution-based counterparts when used in VLMs: achieving 3.2$\times$ faster TTFT and 3.6$\times$ smaller size than SigLIP-SO400M~\citep{zhai2023siglip}, and 2.3$\times$ faster TTFT and 1.7$\times$ smaller size than ConvNeXT~\citep{ge2024convllava}. We further demonstrate that FastVLM scales effectively as more visual instruction tuning data becomes available.
    \item We systematically study the VLM accuracy-latency trade-off by considering both the vision backbone latency and the LLM prefilling time on actual hardware benchmarks. Our results demonstrate an improved resolution-latency-accuracy trade-off achieved by FastVLM, measured on-device rather than estimates.
\end{itemize}

%% file: sec/2_related_works.tex
\section{Related Works}
\label{sec:related_works}

\textbf{Large Multimodal Models.}
With the emergence of large language models~\cite{touvron2023llama, qwen2, qwen2.5, zheng2023judging, OpenAI_GPT4_2023} and large pretrained vision models, such as CLIP~\cite{CLIP}, trained on web-scale image-text datasets, several multimodal architectures have been proposed to encode images aligned with a large language model (LLM) to enable the interpretation of visual signals. Earlier works like Frozen~\cite{tsimpoukelli2021multimodal} and Florence~\cite{flamingo, awadalla2023openflamingo} used a cross-attention mechanism where the image embeddings are fused with text embeddings in intermediate layers of the LLM. More recently, auto-regressive architectures have gained popularity where the image embedding is fed alongside text as input to an LLM. Some prominent works that use this architecture are LLaVA~\cite{liu2023llava, liu2023improvedllava, liu2024llavanext}, mPLUG-Owl~\cite{ye2023mplugowl, ye2023mplugowl2, ye2024mplugowl3longimagesequenceunderstanding}, InstructBLIP~\cite{instructblip}, BLIP-3~\cite{blip3}, SPHINX~\cite{lin2023sphinx}, MiniGPT-4~\cite{zhu2023minigpt}, VILA~\cite{lin2023vila}, MM1~\cite{mckinzie2024mm1methodsanalysis}, Qwen-VL~\cite{Qwen-VL}, InternVL~\cite{chen2023internvl, chen2024far} and Cambrian-1~\cite{tong2024cambrian1}. Recently, Fuyu~\cite{fuyu-8b} and EVE~\cite{diao2024EVE} introduced a simplified architecture that passes raw images directly to the LLM decoder. Chameleon~\cite{chameleonteam2024chameleonmixedmodalearlyfusionfoundation} introduced early fusion mixed-modal models where images are tokenized using a pretrained codebook. While skipping the image encoder is an intriguing approach, the performance of this new class of models lags behind architectures that use a pretrained image encoder.

\textbf{Efficient Image Encoding.}
CLIP~\cite{CLIP} pretrained vision transformers~\cite{ViT} are widely used for encoding images in vision-language models, with popular choices including SigLIP~\cite{zhai2023siglip}, EVA-CLIP~\cite{EVA-CLIP}, InternViT~\cite{chen2023internvl} and DFN-CLIP~\cite{DFN}. To enhance performance, recent works~\cite{karamcheti2024prismatic, tong2024cambrian1, shi2024eagle} employ ensembles of vision encoders trained with different objectives. These works are orthogonal to our work as they can benefit from using an efficient vision encoder among the ensemble of vision encoders. Since ViT-based architectures are a popular choice for VLMs, inefficiencies arise from the number of visual tokens outputted by the encoder, prompting methods like LLaVA-PruMerge~\cite{shang2024LLaVA-PruMerge} and Matryoshka-based token sampling~\cite{hu2024matryoshka, cai2024matryoshka} to dynamically prune tokens. Other approaches~\cite{instructblip, cha2023honeybee, chu2023mobilevlm, chu2024mobilevlm} reduce tokens using perceiver-style resamplers or pooling techniques. Rather than using an isotropic architecture like ViT and then designing custom resamplers and projectors, hierarchical architectures can be a simpler design choice. Hierarchical backbones like ConvNeXT~\cite{liu2022convnet} and FastViT~\cite{vasu2023fastvit} produce fewer tokens as they downsample the input tensor at every stage of compute. Recently, ConvLLaVA~\cite{ge2024convllava} was introduced that uses a pure-convolutional vision encoder to encode images for a VLM. In our work, we introduce an improved convolution-transformer hybrid architecture for VLMs and discuss the Pareto optimal operating points when this architecture is scaled to higher input resolutions.

%% file: sec/3_method.tex
\section{Architecture}\label{sec:architecture}

\begin{figure*}[t]
    \centering
    \includegraphics[width=1.0\linewidth]{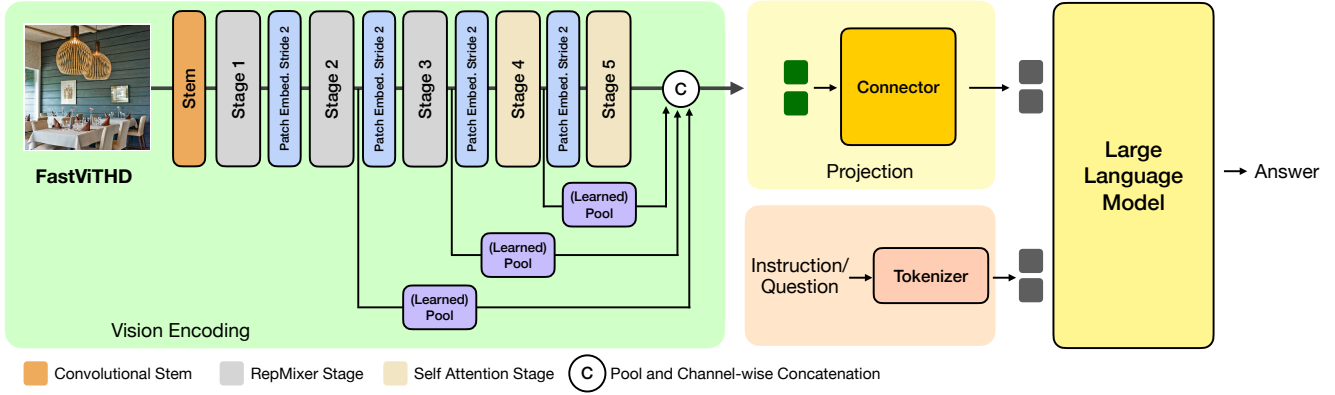}
    \vspace*{-20pt}
    \caption{\textbf{Overview of the FastVLM architecture.} FastVLM consists of our novel vision encoder, \oursImgEnc{}, trained using the same setup as LLaVA. The \oursImgEnc{} architecture is designed for low latency at high resolution, by utilizing additional self-attention layers, and downsampling to generate 4$\times$ fewer tokens than FastViT, and 16$\times$ fewer tokens than ViT-L/14 at resolution 336. }
    \label{fig:arch}
    \vspace*{-5pt}
\end{figure*}

In this section, we first explore the adoption of the FastViT hybrid vision encoder for vision-language modeling. We then introduce architectural interventions to improve performance on VLM tasks. We present~\oursImgEnc{}, a new hybrid vision encoder designed for an efficient high-resolution VLM. We provide comprehensive ablations to demonstrate the optimality of \oursImgEnc{} over FastViT and prior works for different LLMs and input resolutions.  \Cref{fig:arch} illustrates the overall architecture of FastVLM and \oursImgEnc{}. The training setup for all results in this section follows the same configuration as LLaVA-1.5~\citep{liu2023improvedllava} with Vicuna-7B~\cite{zheng2023judging} as the LLM decoder, unless mentioned otherwise. See \cref{sec:experiments} for more details.

\subsection{FastViT as VLM Image Encoder}
\label{sec:fastvit}
VLMs such as LLaVA have three main components: an image encoder, a vision-language projector, and a large language model (LLM). Both the performance and runtime efficiency of a VLM highly depend on its vision backbone. Encoding images at high resolution is essential for achieving strong performance across various VLM benchmarks, especially for text-rich tasks. Therefore, a vision encoder with scalable resolution is particularly beneficial for VLMs. We identify hybrid vision encoders (convolutional layers followed by transformer blocks) as an ideal candidate for VLMs, as their convolutional component enables native resolution scaling, and their transformer blocks further refine high-quality visual tokens for consumption by the LLM.

We use a CLIP-pretrained hybrid vision encoder, specifically the MCi2 image encoder from MobileCLIP~\citep{mobileclip2024}, which has 35.7M parameters and is based on the FastViT architecture. For simplicity, we refer to this encoder as ``FastViT'' throughout the rest of the paper.
As shown in \cref{tab:fastvit_as_enc}, using FastViT at its CLIP-pretrained resolution (256$\times$256) alone does not yield a strong VLM. The main advantage of a hybrid encoder like FastViT lies in its favorable image resolution scaling characteristics, meaning it generates 5.2$\times$ fewer tokens than the ViT architecture with a patch size of 14. The token reduction gives significant advantage to VLMs, as it reduces the prefilling time and time-to-first-token (TTFT) of the transformer decoders.
When the input resolution of FastViT is scaled to 768$\times$768, it produces the same number of visual tokens as ViT-L/14 with an input resolution of 336$\times$336 but achieves better performance on VLM benchmarks. This performance gap is even more pronounced on text-rich benchmarks like TextVQA and DocVQA, despite both architectures producing the same number of visual tokens. Moreover, even with the same token count at higher resolution, it encodes images much faster due to efficient convolution layers.

\begin{table}[t]
\resizebox{0.99\columnwidth}{!}{%
    \setlength{\tabcolsep}{2pt}
    \begin{tabular}{lccccccccc}
    \toprule
    Image  & Input & \#Visual & Latency & \multirow{2}{*}{GQA}  & \multirow{2}{*}{TextVQA} & \multirow{2}{*}{POPE} & \multirow{2}{*}{DocVQA} & Seed & \multirow{2}{*}{Avg-5}\\
    Encoder& Res.  &  Tokens  & Enc.(ms)$\downarrow$ &    &         &      &        &   Bench$^I$  &      \\
    \midrule
    ViT-L/14      & 336        & 576        &  127.4   & 62.0 & 58.2    & 85.9 & 28.1   & 66.1      & 60.1 \\
    ViT-L/14\textsuperscript{\textdagger}      & 336        & 576     &   127.4     & 63.5 & 59.2    & 86.3 & 28.7   & 68.6      & 61.2 \\
    \midrule
    FastViT     & 256        & 64       &   3.0    & 60.2 & 51.6    & 82.9 & 15.8   & 61.5      & 54.4 \\
    FastViT     & 768        & 576      &   34.5    & 62.7 & 62.3    & 86.5 & 34.4   & 67.1      & \textbf{62.6} \\
    \bottomrule
    \end{tabular}%
}
\caption{
\textbf{FastViT has higher accuracy than ViT-L/14 at near 4$\times$ lower latency.}
To scale resolution up to 768, FastViT is made trainable during Stage-2 training of LLaVA-1.5 setup. \textdagger To have a fair comparison, we also report the performance of ViT-L/14 finetuned during Stage-2 training of LLaVA-1.5. All latencies are reported in milliseconds. See \cref{sec:experiments} for details.
}\label{tab:fastvit_as_enc}
    \vspace*{-5pt}
\end{table}

\vspace*{-15pt}
\begin{table}[t]
\resizebox{0.99\columnwidth}{!}{%
    \setlength{\tabcolsep}{2pt}
    \begin{tabular}{lcccccccc}
    \toprule
    Image  & Multi  & Pool & \multirow{2}{*}{GQA}  & \multirow{2}{*}{TextVQA} & \multirow{2}{*}{POPE} & \multirow{2}{*}{DocVQA} & Seed & \multirow{2}{*}{Avg-5}\\
    Encoder& Scale  &  Type  &      &         &      &        &   Bench$^I$  &      \\
    \midrule
    FastViT     &            & -            & 62.7 & 62.3    & 86.5 & 34.4   & 67.1      & 62.6 \\
    FastViT     & \checkmark & AvgPool      & 63.0 & 62.2    & 86.2 & 35.1   & 66.9      & 62.7 \\
    FastViT     & \checkmark & DWConv       & 63.0 & 62.5    & 86.8 & 34.7   & 67.4      & 62.9 \\
    \bottomrule
    \end{tabular}%
}
\caption{
\textbf{Pushing FastViT VLM performance using multi-scale features and pooling strategies.}
These modifications slightly improve FastViT. Training setup is LLaVA-1.5 with Vicuna 7B.
}\label{tab:multiscale_features}
    \vspace*{-10pt}
\end{table}

\vspace*{10pt}
\subsubsection{Multi-Scale Features}\label{sec:multi-scale}
Typical convolutional and hybrid architectures split up the computations into 4 distinct stages with a downsampling operation between them. While the VLM relies on features from the penultimate layer, features in earlier stages of the network extract information at different granularity. Aggregating information from multiple scales can complement high-level features from the penultimate layer.

The architecture for multiple scale feature extraction is shown in \cref{fig:arch}. We ablate between 2 designs to pool features from different stages, i.e. AvgPooling and 2D Depthwise convolutions. From \cref{tab:multiscale_features}, we find that using depthwise convolutions results in better performance. 

\begin{figure}[ht]
    \centering
    \includegraphics[width=0.4\textwidth]{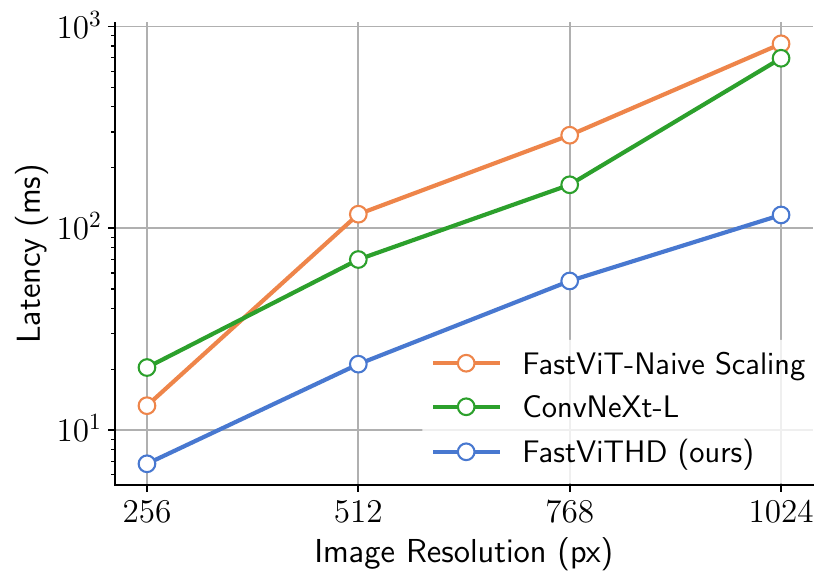}
    \vspace*{-5pt}
    \caption{
    \textbf{Novel scaling strategy of \oursImgEnc{} lowers latency
    at various image resolutions.}
    FastViT-Naive, a naive scaling of the FastViT architecture, and our proposed \oursImgEnc{} have the same number of parameters. ConvNeXt-L is provided for reference. All models are benchmarked on M1 Macbook Pro and trained with LLaVA-1.5 setup and Vicuna 7B. Note that the $y$-axis is in log scale.}
    \label{fig:res_scaling}
    \vspace*{-5pt}
\end{figure}

\subsection{\oursImgEnc{}: High Resolution Encoder for VLM} \label{sec:new-arch}
While FastViT with the introduced model interventions performs well as an image encoder that is 8.7$\times$ smaller than ViT-L/14, 
previous studies~\cite{chen2023internvl, li2024inferenceoptimalvlmsneed} have demonstrated that increasing the scale of the image encoder improves its generalization capabilities.

Hybrid architectures~\cite{moat, yu2024metaformer} typically scale the number of self-attention layers and width in a 4-stage design, but this has drawbacks. From \cref{fig:res_scaling}, simply scaling-up the number of self-attention layers in stages 3 and 4 of FastViT, as done in prior works, is suboptimal and slower than ConvNeXT-L. To mitigate this, we introduce an extra stage with a downsampling layer, ensuring self-attention operates on tensors downsampled by a factor of 32, rather than 16 as in recent models like ViTamin, see \cref{fig:arch}. More details on the naive scaling approach can be found in Sec.\hyperlink{supp_arc_details}{\ref*{sec:arc_ablation}}.
Our design reduces image encoding latency and generates 4$\times$ fewer tokens for the compute-intensive LLM decoder, thereby decreasing the time-to-first-token (TTFT). The architecture schematic is shown in \cref{fig:arch}, and we call this model \textbf{\oursImgEnc{}}.

The model architecture consists of 5 stages, as shown in \cref{fig:arch}, with the first three stages utilizing RepMixer~\cite{vasu2023fastvit} blocks and the last two stages employing multi-headed self-attention~\cite{ViT} blocks. The model depth at each stage is \texttt{[2, 12, 24, 4, 2]}, and the embedding dimensions for each stage are \texttt{[96, 192, 384, 768, 1536]}. The MLP expansion ratio for the ConvFFN layers is set to 4.0. The model has 125.1M parameters, which is 3.5$\times$ larger than the largest FastViT variant from MobileCLIP, but is still smaller than popular ViT alternatives.

\begin{table}[t]
\resizebox{0.99\columnwidth}{!}{%
    \begin{tabular}{l|ccc|ccc}
    \toprule
    Image  & Encoder & Input & Latency  & Zero-Shot  & Avg Perf. & Avg Perf.   \\
    Encoder & Size(M)$\downarrow$ & Res. & Enc.(ms)$\downarrow$  &  ImageNet  & Retrieval  &   on 38 tasks    \\
    \midrule

    ViT-L/14~\cite{DataComp} & 304 & 224 & 47.2 & 79.2 & 60.8 & 66.3 \\
    ViTamin-L~\cite{chen2024vitamin} & 333 & 224 & 38.1 & 80.8 & 60.3 & 66.7 \\
    ConvNeXt-L & 200 & 320 & 34.4 & 76.8 & 64.8 & 63.9 \\
    \textbf{\oursImgEnc{}} & 125 & 224 & 6.8 & 78.3 & 67.7 & 66.3 \\

    \bottomrule
    \end{tabular}%
}
\caption{
\textbf{\oursImgEnc{} achieves competitive results on CLIP benchmarks at significantly lower latency.} We follow the same setup described in~\cite{chen2024vitamin} to report average retrieval performance and setup described in~\cite{DataComp} to report average performance on 38 tasks. All models are benchmarked on M1 Macbook Pro.
}\label{tab:clip_results}
    \vspace*{-10pt}
\end{table}

We follow the CLIP pretraining setup of \cite{mobileclip2024} using the DataCompDR-1B dataset to pretrain~\oursImgEnc{} before employing it for FastVLM training. \Cref{tab:clip_results} shows that~\oursImgEnc{}, despite being 2.4$\times$ smaller and 6.9$\times$ faster than ViT-L/14, achieves comparable average performance across 38 multi-modal zero-shot tasks~\cite{DataComp}. In comparison to ViTamin~\cite{chen2024vitamin}, a hybrid transformer architecture built for VLMs, \oursImgEnc{} delivers superior average retrieval performance while being 2.7$\times$ smaller and 5.6$\times$ faster. In \cref{tab:backbone_ablation}, we compare \oursImgEnc{} with other CLIP-pretrained hierarchical backbones, i.e. ConvNeXT-L and ConvNeXT-XXL, for VLM tasks after LLaVA-1.5 training. \oursImgEnc{} performs as well as ConvNeXT-XXL while being 6.8$\times$ smaller and 3.3$\times$ faster.

\begin{table}[t]
\resizebox{0.99\columnwidth}{!}{%
    \setlength{\tabcolsep}{2pt}
    \begin{tabular}{lccccccccc}
    \toprule
    Image  & Input & Latency & \#Visual & \multirow{2}{*}{GQA}  & \multirow{2}{*}{TextVQA} & \multirow{2}{*}{POPE} & \multirow{2}{*}{DocVQA} & Seed & \multirow{2}{*}{Avg-5} \\
    Encoder& Res. & Enc.(ms)$\downarrow$  &  Tokens  &      &         &      &        &   Bench$^I$  &       \\
    \midrule
    \textbf{\oursImgEnc{}} & 256 & 10.1 & 16 & 60.6 & 53.1 & 82.3 & 17.4 & 63.7 & 55.5 \\

    \midrule
    C.N-L   & 320 & 34.4 & 100 & 61.9 & 55.5 & 85.3 & 21.3 & 64.6 & 57.7 \\
    C.N-XXL & 256 &  89.9 & 64 & 62.7 & 56.3 & 85.3 & 21.6 & 65.6 & 58.3 \\
    \textbf{\oursImgEnc{}} & 512  & 33.5 & 64 & 63.0 & 59.3 & 86.4 & 25.7 & 67.1 & \textbf{60.4} \\
    
    \midrule

    \textbf{\oursImgEnc{}}     & 768    & 122.6   & 144 & 62.4 & 62.9 & 87.7 & 32.9  & 68.2 & 62.8 \\

    \midrule

    C.N-L         & 512 &  71.9     & 256 & 61.8 & 61.0 & 86.3 & 30.8 & 66.8 & 61.3 \\
    C.N-XXL       & 512 &   397.1   & 256 & 62.3 & 65.1 & 87.7 & 36.2 & 68.4 & \textbf{63.9} \\
    \textbf{\oursImgEnc{}}     & 1024 & 235.6    & 256 &  63.1 & 64.4 & 88.1 & 35.6 & 68.5 & \textbf{63.9} \\

    \bottomrule
    \end{tabular}%
}
\caption{
\textbf{\oursImgEnc{} achieves higher accuracy than ConvNeXT while having lower latency at a higher resolution.}
The models are grouped based on the total number of visual tokens produced for the LLM to process. ``C.N'' stands for ConvNeXT. Training setup is LLaVA-1.5 with Vicuna 7B. 
}\label{tab:backbone_ablation}
    \vspace*{-5pt}
\end{table}

\subsubsection{Vision Encoder - Language Decoder Interplay}

\begin{figure}[t]
    \centering
    \includegraphics[width=0.42\textwidth]{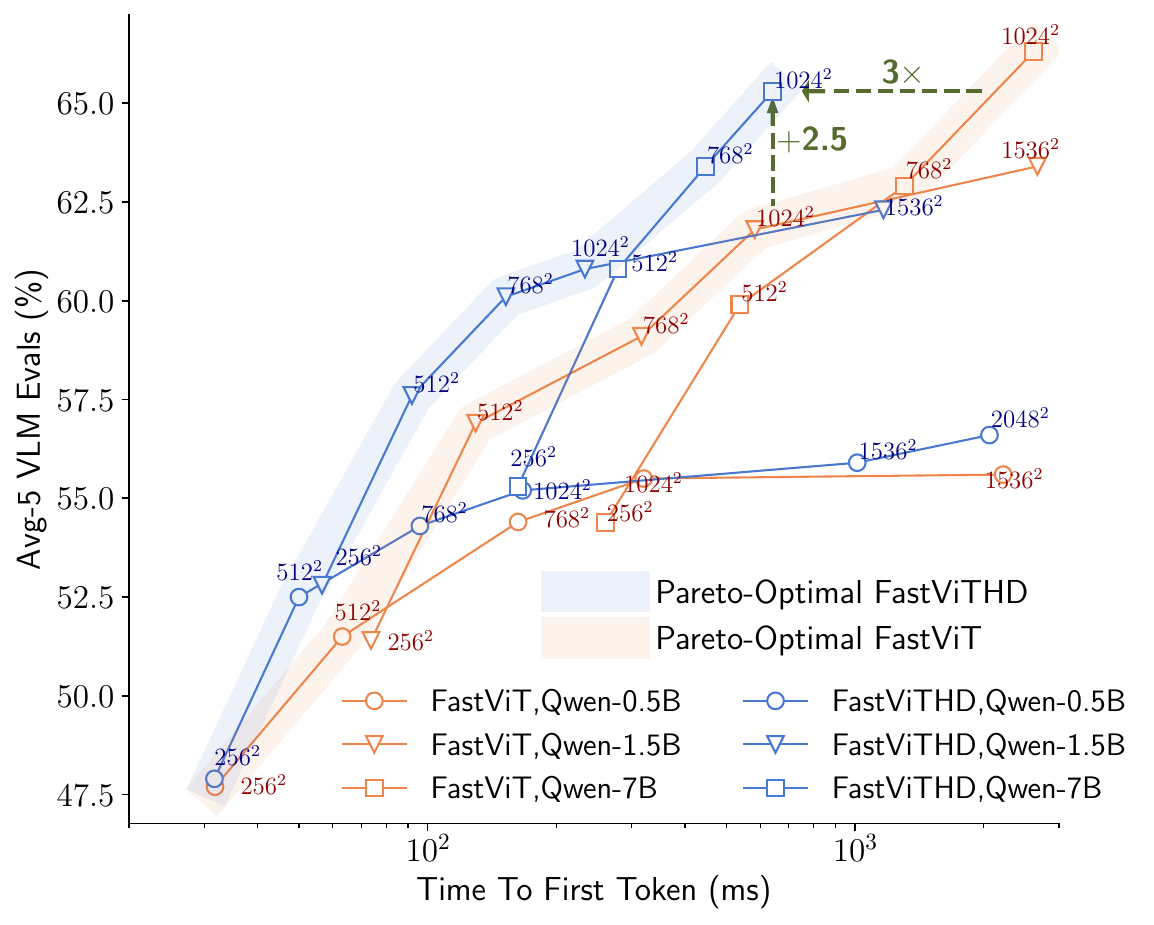}
    \vspace*{-5pt}
    \caption{
    \textbf{\oursImgEnc{} improves the Pareto-Optimal curve for accuracy versus time to first token compared with FastViT.}
    Comparison of FastViT and \oursImgEnc{} backbones paired with Qwen2~\citep{qwen2} family (chat variant) LLMs of varying sizes and different image resolutions (annotated for each point). The Pareto-optimal curve is highlighted for the two vision backbones. Training setup is LLaVA-1.5. Note that the $x$-axis is in log scale.}
    \label{fig:dec_enc}
    \vspace*{-5pt}
\end{figure}

\begin{figure}[t]
    \centering
    \includegraphics[width=0.35\textwidth]{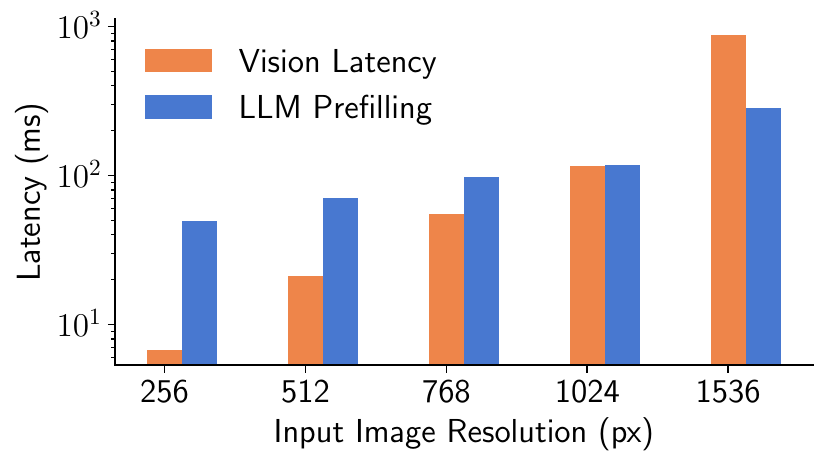}
    \vspace*{-5pt}
    \caption{
    \textbf{Vision latency dominates at high resolution.}
    Breakdown of FastVLM's time to first token for varying image resolutions. Vision encoder is \oursImgEnc{} and LLM is Qwen2-1.5B.}
    \label{fig:ttft_breakdown}
    \vspace*{-5pt}
\end{figure}

\begin{figure}[ht]
    \centering
    \includegraphics[width=0.4\textwidth]{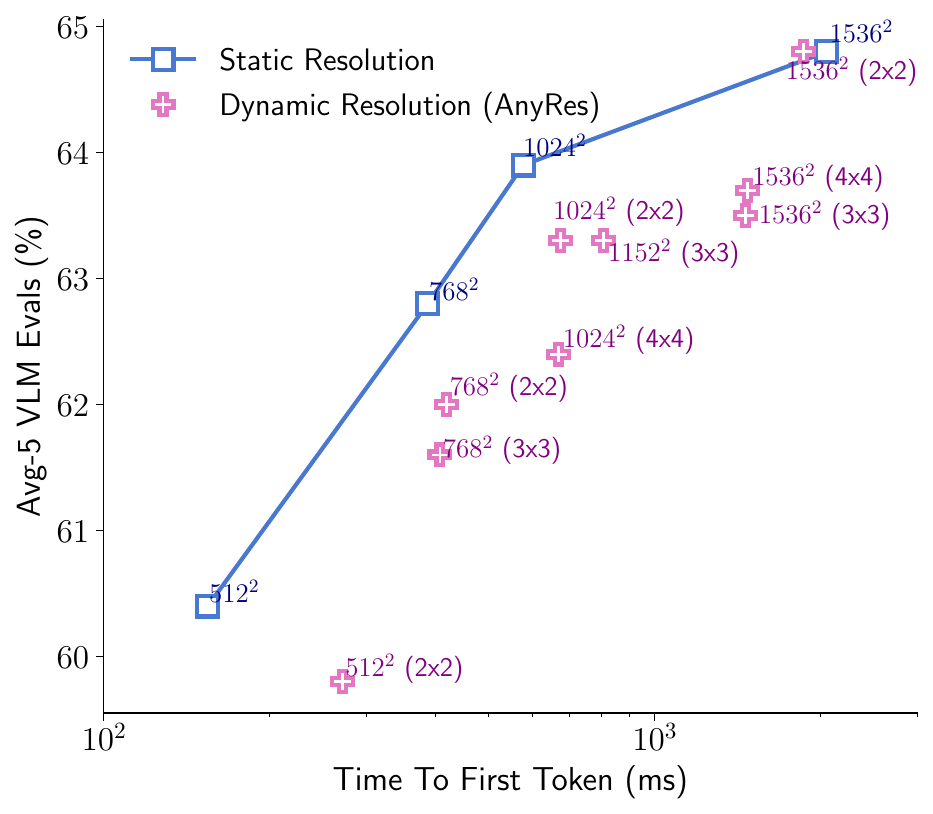}
    \vspace*{-5pt}
    \caption{\textbf{Dynamic input resolution (AnyRes) is only optimal at the highest resolution when using fewer tiles (2$\times$2).} The vision encoder is \oursImgEnc{}. The tile grid size is specified in parenthesis. Training setup is LLaVA-1.5 with Vicuna 7B. Note that the $x$-axis is in log scale.}
    \label{fig:res_compare}
    \vspace*{-5pt}
\end{figure}

The accuracy-latency trade-off in a VLM is influenced by several factors. On one hand, the overall performance of the VLM depends on (1) the input image resolution, (2) the quantity and quality of visual tokens, and (3) the capability of the LLM. On the other hand, the total latency (time to first token generation) of a VLM is determined by (1) the latency of the vision encoder and (2) the prefilling time of the LLM. The latter is affected by both the number of tokens produced by the vision encoder and the size of the LLM.

Due to the complex optimization landscape of VLMs, claims regarding the optimality of a vision encoder must be verified across various pairs of (Resolution, LLM). Here, we empirically demonstrate the optimality of \oursImgEnc{} over FastViT. For each vision encoder, we consider three LLMs, Qwen2~\citep{qwen2}-0.5B/1.5B/7B, along with a range of input image resolutions. For each (Resolution, LLM) pair, we conduct LLaVA-1.5~\citep{liu2023improvedllava} pretraining and visual instruction tuning, and evaluate the resulting model over a range of tasks. The results are presented in \cref{fig:dec_enc}.

First, we observe that for a vision encoder, the Pareto-optimal curve (highlighted in \cref{fig:dec_enc}), which represents the maximum achievable performance for a given runtime budget (TTFT), consists of varying sizes of LLMs.
Specifically, pairing high resolution with a small LLM is suboptimal as
a small LLM cannot effectively utilize that many tokens, and TTFT will be dominated by the latency of the vision encoder (see \cref{fig:ttft_breakdown}).

Second, the Pareto-optimal curve for \oursImgEnc{} in \cref{fig:dec_enc} is significantly better than that of FastViT. For a given runtime budget, considering all possible (Resolution, LLM) pairs, we achieve significantly better performance (an improvement of over 2.5 points on the Average-5 metric) with \oursImgEnc{}. Similarly, \oursImgEnc{} can reach a target VLM performance up to 3$\times$ faster. It is important to note that in previous sections, we demonstrated that a FastViT-based VLM already represents a significant improvement over ViT-based VLMs, and yet \oursImgEnc{} provides substantial gains over FastViT.

\subsubsection{Static vs. Dynamic Input Resolution}

There are two ways to scale input resolution: adjusting the model’s input resolution directly or tiling the image and setting the encoder’s resolution to the tile size.
The tiled inference (AnyRes) was introduced in prior works \cite{lin2023sphinx, liu2023llava} to enable ViT models to process high resolution images. Since \oursImgEnc{} is designed to run inference efficiently on high input resolutions, we analyze the optimal operating point for various resolutions using the two strategies. 
From \cref{fig:res_compare}, we see that setting the model's input resolution directly to the desired resolution offers the best accuracy-latency tradeoff, with dynamic resolution benefiting only at extreme resolutions like 1536$\times$1536, due to memory bandwidth limitations.
If dynamic resolution is desired, using a setting with fewer tiles exhibits better accuracy-latency tradeoff. Further discussion on this setup is presented in Sec. \hyperlink{supp_anyres_results}{\ref*{sec:anyres_results}}.

\subsubsection{Comparison with Token Pruning \& Downsampling}

We further compare the performance of \oursImgEnc{} operating at different resolutions to popular token pruning methods in literature. From \cref{tab:comp_token_pruning}, we find that VLMs achieve better accuracy to latency trade-off using a hierarchical backbone as opposed to using token pruning methods on isotropic architectures like ViT. By simply training the VLMs at lower input resolution, \oursImgEnc{}  achieves visual token counts as low as 16, while improving over recent token pruning methods. Interestingly, even the most effective token pruning methods, such as those proposed by~\cite{cai2024matryoshka, hu2024matryoshka, huang2024dynamicllava, yang2024visionzip}, perform worse than~\oursImgEnc{} trained at a lower input resolution of 256$\times$256.

\begin{table}[t]
\resizebox{0.99\columnwidth}{!}{%
    \setlength{\tabcolsep}{2pt}
    \begin{tabular}{lcccccccc}
    \toprule
    \multirow{2}{*}{Model}  & Input & \#Visual & \multirow{2}{*}{GQA}  & \multirow{2}{*}{SQA} & Text- & \multirow{2}{*}{POPE} & VQA & Seed \\
    & Res. &  Tokens  &      &     &  VQA  &      &  v2   &   Bench \\
    \midrule
    ViT-L/14 M$^3$~\cite{cai2024matryoshka} & 336 & 9 & 58.0 & - & - & \textbf{83.4} & -  & 55.4 \\
    ViT-L/14 MQT~\cite{hu2024matryoshka} & 336 & 16 & 57.6 & 67.5 & - & 80.8 &  71.1 & - \\
    \textbf{\oursImgEnc{}} & 256 & 16 & \textbf{60.6} & \textbf{69.2} & 53.1 & 82.3 & \textbf{74.7} & \textbf{58.8} \\

    \midrule
    ViT-L/14 PruMerge~\cite{shang2024LLaVA-PruMerge} & 336 & 40 & - & 68.5 & 56.0 & 76.3 & 72.0 & - \\
    ViT-L/14 PruMerge+~\cite{shang2024LLaVA-PruMerge} & 336 & 40 & - & 68.3 & 57.1 & 84.0 & 76.8 & - \\
    ViT-L/14 M$^3$~\cite{cai2024matryoshka} & 336 & 36 & 60.3 & - & - & 85.5 & - & 58.0 \\

    FastV~\cite{chen2024fastv}              & 336 & 64 & 46.1 & 51.1 & 47.8 & 48.0 & 55.0 & 51.9 \\
    SparseVLM~\cite{zhang2024sparsevlm}     & 336 & 64 & 52.7 & 62.2 & 51.8 & 75.1 & 68.2 & 51.1 \\
    VisionZip~\cite{yang2024visionzip}      & 336 & 64 & 55.1 & 69.0 & 55.5 & 77.0 & 62.9 & 52.2 \\
    VisionZip\ddag~\cite{yang2024visionzip} & 336 & 64 & 57.0 & 68.8 & 56.0 & 80.9 & 74.2 & 53.4 \\

    DynamicLLaVA$_{I}$~\cite{huang2024dynamicllava} & 336 & 115 & 61.4 & \textbf{69.1} & 57.0 & 85.0 & \textbf{78.0} & - \\

    DynamicLLaVA$_{I|T}$~\cite{huang2024dynamicllava} & 336 & 115 & 61.3 & 68.6 & 56.5 & 85.9 & 77.9 & - \\
    
    \textbf{\oursImgEnc{}} & 512  & 64 & \textbf{63.0} & 68.9 & \textbf{59.3} & \textbf{86.4} & \textbf{78.0} & \textbf{61.8} \\
    
    \midrule

    ViT-L/14 M$^3$~\cite{cai2024matryoshka} & 336 & 144 & 61.3 & - & - & 87.0 & - & 59.7 \\
    ViT-L/14 MQT~\cite{hu2024matryoshka} & 336 & 144 & 61.4 & 67.6 & - & 83.9 & 76.4 & - \\

    FastV~\cite{chen2024fastv}              & 336 & 192 & 52.7 & 67.3 & 52.5 & 64.8 & 67.1 & 57.1 \\
    SparseVLM~\cite{zhang2024sparsevlm}     & 336 & 192 & 57.6 & \textbf{69.1} & 56.1 & 83.6 & 75.6 & 55.8 \\
    VisionZip~\cite{yang2024visionzip}      & 336 & 192 & 59.3 & 68.9 & 57.3 & 85.3 & 76.8 & 56.4 \\
    VisionZip\ddag~\cite{yang2024visionzip} & 336 & 192 & 60.1 & 68.2 & 57.8 & 84.9 & 77.4 & 57.1 \\
    
    \textbf{\oursImgEnc{}}     & 768    & 144   & \textbf{62.4} & 67.6 & \textbf{62.9} & \textbf{87.7} & \textbf{78.9} & \textbf{62.5} \\

    \midrule

    ViT-L/14 MQT~\cite{hu2024matryoshka} & 336 & 256 & 61.6 & \textbf{67.5} & - & 84.4 & 76.8 &  - \\
    \textbf{\oursImgEnc{}}     & 1024 & 256    & \textbf{63.1} & 67.4 & 64.4 & \textbf{88.1} & \textbf{79.2}  & - \\

    \bottomrule
    \end{tabular}%
}
\caption{
\textbf{\oursImgEnc{} more effectively reduces tokens compared with token pruning methods.}
The models are grouped based on total number of visual tokens. ``-'' indicates that performance was not reported in the respective paper. All models presented in this table are trained using LLaVA-1.5 setup with Vicuna 7B. \ddag - indicates further finetuning as reported in~\cite{yang2024visionzip}. $_{I}$ - indicates vision only sparsification and $_{I|T}$ indicates vision-language sparsification, as reported in~\cite{huang2024dynamicllava}.
}\label{tab:comp_token_pruning}
\end{table}

%% file: sec/4_experiments.tex

\definecolor{mintbg}{rgb}{.63,.79,.95}
\colorlet{lightmintbg}{mintbg!30}
\definecolor{pumpkin}{rgb}{0.75, 0.75, 0.75}
\colorlet{lightpumpkin}{pumpkin!20}

\begin{table*}[t]
\resizebox{2.0\columnwidth}{!}{
    \setlength{\tabcolsep}{2pt}
    \begin{tabular}{l|cccc|cc|cc|cccccccccc}
    \toprule
    Row & \multirow{2}{*}{Method} & Vision & \multirow{2}{*}{LLM} & Data (M) & Input & \#Visual & Vis. Enc. & TTFT & \multirow{2}{*}{GQA}  & \multirow{2}{*}{SQA} & Text & \multirow{2}{*}{POPE} & LLaVA & MM- & VQA & Doc & \multirow{2}{*}{MMMU} & Seed \\
    Ann. & & Encoder & & (PT+IT) & Res. &  Tokens & Size(M)$\downarrow$  & (ms)$\downarrow$  &  &   & VQA  &      &  Bench$^W$   & Vet & v2 & VQA & &  Bench$^I$      \\
    \midrule
    \multicolumn{19}{c}{0.5B Model Comparison} \\
    \midrule

    R1 & nanoLLaVA & ViT-SO400M & Qw.1.5 & - & 384 & 729 & 430 & 535 &  54.8 & 59.0 & 46.7 & 84.1 & - & - & 70.8 & - & 30.4 & - \\

    R2 & LLaVAOV~\cite{li2024llava}$^*$ & ViT-SO400M & Qw.2 & 4.5+3.2 & 1152 & 7290 & 430 & 14124 &  - & 67.2 & - & - & - & 29.1 & - & 70.0 & 31.4 & 65.5 \\ 

    \rowcolor{lightmintbg} R3 & \textbf{FastVLM (Ours)} & \oursImgEnc{} & Qw.2 & 15+1.1 & 1024 & 256 & \textbf{125} & \textbf{166} & 61.6 & 61.4 & 57.4 & \textbf{87.4} & 56.0 & 31.8 & 77.0 & 61.0 & 30.9 & 65.6 \\

    \rowcolor{lightmintbg} R4 & \textbf{FastVLM (Ours)} & \oursImgEnc{} & Qw.2 & 15+12.5 & 1024 & 256 & \textbf{125} & \textbf{166} & \textbf{63.1} & 81.5 & 62.9 & 86.6 & 66.7 & 29.8 & \textbf{78.8} & 70.4 & 32.9 & 69.2 \\

    \rowcolor{lightmintbg} R5 & \textbf{FastVLM (Ours)} & \oursImgEnc{} & Qw.2 & 15+23.1 & 1024 & 256 & \textbf{125} & \textbf{166} & 62.7 & \textbf{82.0} & \textbf{65.8} & 86.2 & - & \textbf{37.5} & 78.6 & \textbf{79.1}  & \textbf{33.6} & \textbf{69.3} \\

    \midrule
    \multicolumn{19}{c}{1-2B Model Comparison} \\
    \midrule

    R6 & MobileVLMv2~\cite{chu2024mobilevlm}& ViT-L/14 & ML. & 1.2+3.6 & 336 & 144 & 304 & 458 & 59.3 & 66.7 & 52.1 & 84.3 & - & - & - & - & - & - \\

    \rowcolor{lightmintbg} R7 & \textbf{FastVLM (Ours)} & \oursImgEnc{} & Qw.2 & 15+1.1 & 768 & 144 & \textbf{125} & \textbf{152}  & 63.9 & 75.8 & 64.4 & 87.2 & 65.2 & 35.4 & 79.4 & 61.3 &  34.9 & 71.7 \\

    \rowcolor{lightmintbg} R8 & \textbf{FastVLM (Ours)} & \oursImgEnc{} & Qw.2 & 15+12.5 & 768 & 144 & \textbf{125} & \textbf{152}  & \textbf{64.2} & \textbf{87.9} & \textbf{66.4} & \textbf{88.3} & \textbf{68.5} & \textbf{41.1} & \textbf{80.1} & \textbf{69.4} & \textbf{38.1} & \textbf{72.4} \\

    \midrule

    R9 & DeepSeekVL~\cite{lu2024deepseekvl} & ViT-SO400M & DS. & - & 384 & 576 & 430 & - & - & - & - & 87.6 & - & 34.8 & - & - & 32.2 & 66.7 \\

    R10 & MM1~\cite{mckinzie2024mm1methodsanalysis}$^*$ & ViT-H & - & 3000+1.5 & 1344 & 720 & 632 & - &  
    - & 62.3 & 68.2 & 87.4  & 67.5 & 39.4 & - & 68.4 & 33.2 & 65.6 \\

    \rowcolor{lightmintbg} R11 & \textbf{FastVLM (Ours)} & \oursImgEnc{} & Qw.2 & 15+1.1 & 1024 & 256 & \textbf{125} & \textbf{233} & 64.2 & 74.8 & 66.0 & \textbf{88.0} & 66.9 & 37.6 & 79.9 & 67.7 & 33.1 & 71.4 \\

    \rowcolor{lightmintbg} R12 & \textbf{FastVLM (Ours)} & \oursImgEnc{} & Qw.2 & 15+12.5 & 1024 & 256 & \textbf{125} & \textbf{233} & \textbf{64.3} & 89.9 & 69.0 & 87.8 & 68.0 & 43.6 & 80.7 & 75.6 & \textbf{39.1} & \textbf{73.1} \\

    \rowcolor{lightmintbg} R13 & \textbf{FastVLM (Ours)} & \oursImgEnc{} & Qw.2 & 15+23.1 & 1024 & 256 & \textbf{125} & \textbf{233} & \textbf{64.3}  & \textbf{92.9} & \textbf{71.2} &  87.8 & - & \textbf{51.0} & \textbf{80.9}  & \textbf{84.2} & 38.0 & 72.6 \\

    \midrule
    \multicolumn{19}{c}{7B Model Comparison} \\
    \midrule

    R14 & InstructBLIP~\cite{instructblip} & ViT-g/14 & Vic. & 129+1.2 & 224 & 32 & 1012 & 302 & 49.2 & 60.5 & 50.1 & - & 60.9 & 26.2 & - & - & 30.6 & - \\ 
    
    \rowcolor{lightmintbg} R15 & \textbf{FastVLM (Ours)} & \oursImgEnc{} & Vic. & 0.5+0.6 & 256 & 16 & \textbf{125} & \textbf{150} & 60.6 & 69.2 & 53.1 & 82.3 & 60.4 & 27.5 & 74.7 & 17.4 & 36.2 & 63.7 \\

    \rowcolor{lightmintbg} R16 & \textbf{FastVLM (Ours)} & \oursImgEnc{} & Vic. & 15+1.1 & 256 & 16 & \textbf{125} & \textbf{150} & \textbf{62.1} & \textbf{75.7} & \textbf{57.2} & \textbf{83.9} & \textbf{64.0} & \textbf{31.5} & \textbf{77.3} & \textbf{29.8} & \textbf{37.6} & \textbf{68.8} \\

    \midrule

    R17 & MobileVLMv2~\cite{chu2024mobilevlm}& ViT-L/14 & Vic. & 1.2+3.6 & 336 & 144 & 304 & 460 &  62.6 & 74.8 & 62.3 & 85.3 & - & - & - & - & - & - \\

    R18 & ConvLLaVA~\cite{ge2024convllava} & ConvNeXT-L & Vic. & 4.9+0.6 & 768 & 144 & 200 & 496 & - & - & 59.1 & 87.3 & - & 44.8 & - & 44.8 & 36.3 & 68.8 \\

    \rowcolor{lightmintbg} R19 & \textbf{FastVLM (Ours)} & \oursImgEnc{} & Vic. & 0.5+0.6 & 768 & 144 & \textbf{125} & \textbf{387} & 62.4 & 67.6 & 62.9 & \textbf{87.7} & 63.8 & 31.5 & 78.9 & 32.9 & 34.9 & 68.2 \\
    
    \rowcolor{lightmintbg} R20 & \textbf{FastVLM (Ours)} & \oursImgEnc{} & Vic. & 0.5+1.1 & 768 & 144 & \textbf{125} & \textbf{387} & 63.2 & 73.5 & 67.5 & 86.3 & 63.9 & 33.0 & 79.1 & 57.3  & 36.9 & 69.9 \\

    \rowcolor{lightmintbg} R21 & \textbf{FastVLM (Ours)} & \oursImgEnc{} & Vic. & 15+1.1 & 768 & 144 & \textbf{125} & \textbf{387} & 65.0 & 78.7 & 69.4 & 87.5 & 67.0 & 42.2 & 81.3 & 65.5  & 37.0 & 73.7 \\
    
    \rowcolor{lightmintbg} R22 & \textbf{FastVLM (Ours)} & \oursImgEnc{} & Qw.2 & 15+1.1 & 768 & 144 & \textbf{125} & 446 & \textbf{65.6} & 85.9 & 69.5 & 87.2 & 73.0 & 41.3 & \textbf{81.3} & 66.9  & 43.6 & \textbf{75.3} \\

    \rowcolor{lightmintbg} R23 & \textbf{FastVLM (Ours)} & \oursImgEnc{} & Qw.2 & 15+12.5 & 768 & 144 & \textbf{125} & 446 & 64.7 & \textbf{96.0} & \textbf{71.9} & 87.4 & \textbf{73.6} & \textbf{49.4} & 81.1 & \textbf{75.3}  & \textbf{44.6} & 74.3 \\

    \midrule

    R24 & Qwen-VL~\cite{Qwen-VL}      & ViT-G/14 & Qw. & 1400+50 & 448 & 256 & 1844 & - & 59.3 & 67.1 & 63.8 & - & - & - & 79.5 & 65.1 & - & -\\
    
    R25 & Qwen-VL-Chat~\cite{Qwen-VL} & ViT-G/14 & Qw. & 1400+50 & 448 & 256 & 1844 & - & 57.5 & 68.2 & 61.5 & - & - & - & 78.2 & 62.6 & -& -\\

    R26 & ConvLLaVA~\cite{ge2024convllava} & ConvNeXT-L & Vic. & 4.9+0.6 & 1024 & 256 & 200 & 1157 & - & - & 62.5 & 87.7 & - & \textbf{44.4} & - & 48.5 & 35.1 & 69.3 \\

    \rowcolor{lightmintbg} R27 & \textbf{FastVLM (Ours)} & \oursImgEnc{} & Vic. & 0.5+0.6 & 1024 & 256 & \textbf{125} & \textbf{577} & 63.1 & 67.4 & 64.4 & \textbf{88.1} & 64.8 & 31.7 & 79.2 & 35.6 & 35.1 & 68.5 \\
    
    \rowcolor{lightmintbg} R28 &\textbf{FastVLM (Ours)} & \oursImgEnc{} & Vic. & 0.5+1.1 & 1024 & 256 & \textbf{125} & \textbf{577} & 63.3 & 74.1 & 67.4 & 87.1 & 66.5 & 32.4 & 79.3 & 62.8 & 37.3 & 69.9 \\

    \rowcolor{lightmintbg} R29 & \textbf{FastVLM (Ours)} & \oursImgEnc{} & Vic. & 15+1.1 & 1024 & 256 & \textbf{125} & \textbf{577} & 65.2 & 80.3 & 70.6 & 87.2 & 71.5 & 40.1 & 81.6 & 72.4 & 36.7 & 73.5 \\

    \rowcolor{lightmintbg} R30 & \textbf{FastVLM (Ours)} & \oursImgEnc{} & Qw.2 & 15+1.1 & 1024 & 256 & \textbf{125} & 641 & \textbf{65.8} & \textbf{84.9} & \textbf{72.1} & 87.8 & \textbf{75.8} & 44.1 & \textbf{81.7} & \textbf{73.3} & \textbf{46.2} & \textbf{75.1} \\

    \midrule
    
    R31 & LLaVA-1.5~\cite{liu2023improvedllava} & ViT-L/14 &  Vic. & 0.5+0.6 & 336 & 576 & 304 & 1297 & 62.0 & 70.4 & 58.2 & 85.9 & 59.6 & 31.1 & 76.6 & 28.1 & 35.3 & 66.1 \\

    R32 & MobileVLMv2~\cite{chu2024mobilevlm}& ViT-L/14 & Vic. & 1.2+3.6 & 336 & 576 & 304 & 1297 & 64.6 & 74.8 & 66.8 & 86.1 & - & - & - & - & - & - \\

    R33 & ShareGPT4V~\cite{chen2023sharegpt4v} & ViT-L/14 & Vic. & 1.2+0.7 & 336 & 576 & 304 & 1297 & 63.3 & 68.4 & 60.4 & 85.7 & 72.6 & 37.6 & 80.6 & - & - & 69.7 \\

    R34 & ViTamin~\cite{chen2024vitamin} & ViTamin-L & Vic. & 0.5+0.6 & 384 & 576 & 333 & 1308 &  61.6 & 67.6 & 59.8 & 85.5 & 66.1 & 33.6 & 78.9 & - & - & - \\

    R35 & ConvLLaVA~\cite{ge2024convllava} & ConvNeXT-L & Vic. & 4.9+0.6 & 1536 & 576 & 200 & 2740 & - & - & 65.8 & 87.3 & - & 45.9 & - & 59.0 & 35.8 & 70.2 \\

    R36 & VILA~\cite{lin2023vila} & ViT-L/14 & L-2 & 50+1 & 336 & 576 & 304 & 1297 & 62.3 & 68.2 & 64.4 & 85.5 & 69.7 & 34.9 & 79.9 & - & - & 62.8  \\
    
    R37 & LLaVA-FlexAttn~\cite{li2025flexattention} & ViT-L/14 & Vic. & 0.5+0.6 & 1008 & 576 & 304 & - & 62.2 & - & 48.9 & 85.9 & - & 29.4 & 78.7 & - & - & - \\

    R38 & MM1~\cite{mckinzie2024mm1methodsanalysis}$^*$ & ViT-H & - & 3000+1.5 & 1344 & 720 & 632 & - &  
    - & 72.6 & 72.8 & 86.6 & \textbf{81.5} & 42.1 & \textbf{82.8} & 76.8 & 37.0 & 69.9 \\

    R39 & LLaVA-NeXT\textsuperscript{\textdagger}$^*$ & ViT-L/14 & L-3 & - & 672 & 2880 & 304 & 20347 &  
    65.2 & 72.8 & 64.6 & - & 80.1 & - & - & 78.2 & 41.7  & 72.7 \\

    \rowcolor{lightmintbg} R40 & \textbf{FastVLM (Ours)} & \oursImgEnc{} & Qw.2 & 15+6.5 & 1024 & 256 & \textbf{125} & \textbf{641} & \textbf{66.0} & 87.4 & 73.1 & 87.3 & 72.4 & 47.6 & \textbf{82.3} & 78.7 & 42.8 & \textbf{75.9} \\

    \rowcolor{lightmintbg} R41 & \textbf{FastVLM (Ours)} & \oursImgEnc{} & Qw.2 & 15+12.5 & 1024 & 256 & \textbf{125} & \textbf{641} & 65.2 & 95.7 & 73.4 & 86.9 & 71.1 & 48.4 & 81.6 & 82.7 & \textbf{47.3} & 74.1 \\

    \rowcolor{lightmintbg} R42 & \textbf{FastVLM (Ours)} & \oursImgEnc{} & Qw.2 & 15+23.1 & 1024 & 256 & \textbf{125} & \textbf{641} & 64.1 & \textbf{96.8} & \textbf{76.1} & 87.2 & - & \textbf{56.3} & 81.9 & \textbf{88.0} & 44.0 & 74.5 \\

    \midrule
    \multicolumn{19}{c}{VLMs with Multiple Vision Encoders and 8B LLM} \\
    \midrule

     \rowcolor{lightpumpkin} & & ConvNeXT-L &  & & 1536 &   & 200 &  &  &  &  &  &  &  &  & & &  \\
     \rowcolor{lightpumpkin} \multirow{-2}{*}{R43} & \multirow{-2}{*}{MiniGemini-HD\textsuperscript{\textdagger}} & ViT-L/14 & \multirow{-2}{*}{L-3} & \multirow{-2}{*}{1.5+1.5} & 672 & \multirow{-2}{*}{2880} & 304 & \multirow{-2}{*}{21832} &  \multirow{-2}{*}{64.5} & \multirow{-2}{*}{75.1} & \multirow{-2}{*}{70.2} & \multirow{-2}{*}{-} & \multirow{-2}{*}{-} & \multirow{-2}{*}{-} & \multirow{-2}{*}{-} & \multirow{-2}{*}{74.6} & \multirow{-2}{*}{37.3} & \multirow{-2}{*}{73.2} \\

     \hline   
     \rowcolor{lightpumpkin} & & ViT-SO400M &  &  & 384  &   & 430 &  &  &  &  &  &  &  &  & & &  \\
     \rowcolor{lightpumpkin} & & ConvNeXt-XXL &  &  & 1024  &   & 846 &  &  &  &  &  &  &  &  & & &  \\
     \rowcolor{lightpumpkin} & & DINOv2-ViT-L/14 &  &  & 518 &   & 304 &  &  &  &  &  &  &  &  & & &  \\
     \rowcolor{lightpumpkin} 
     \multirow{-4}{*}{R44} & \multirow{-4}{*}{Cambrian-1~\cite{tong2024cambrian1}} & ViT-L/14 & \multirow{-4}{*}{L-3} & \multirow{-4}{*}{2.5+7} & 336 & \multirow{-4}{*}{576} & 304 & \multirow{-4}{*}{5085} & \multirow{-4}{*}{64.6} & \multirow{-4}{*}{80.4}  & \multirow{-4}{*}{71.7} & \multirow{-4}{*}{-} & \multirow{-4}{*}{-} & \multirow{-4}{*}{-} & \multirow{-4}{*}{-} & \multirow{-4}{*}{77.8} & \multirow{-4}{*}{42.7} & \multirow{-4}{*}{74.7} \\

    \bottomrule
    \end{tabular}%
}
\caption{
\textbf{VLM evaluations and comparison with recent methods.} The models are grouped based on total number of visual tokens. ``-'' indicates that performance was not reported in the respective paper. For the dataset column, ``-'' indicates that the dataset size for pretraining (``PT'') or instruction tuning (``IT'') is not explicitly mentioned in the respective paper. For methods that have more than 2 stages of training, we report the total samples used for all the pretraining stages as part of ``PT''. ``TTFT'' means time to first token (the sum of the vision encoder latency and the LLM prefilling time), we report latency only for models that are publicly available and in a format favorable to MLX~\citep{mlx2023} ``Vic.'' refers to Vicuna~\cite{zheng2023judging}, ``Qw.2'' refers to Qwen2~\cite{qwen2} and ``Qw.'' refers to Qwen~\cite{qwen}. ``L-2'' refers to LLaMA-2. ``L-3'' refers to LLaMA-3. ``ML.'' refers to MobileLLaMA~\cite{chu2023mobilevlm, chu2024mobilevlm}. ``DS.'' refers to DeepSeek LLM~\cite{deepseek-llm}. $^*$  For input resolution and visual tokens, we report the highest supported resolution by the respective models as some models like LLaVA-OneVision~\cite{li2024llava} and MM1~\cite{mckinzie2024mm1methodsanalysis} use dynamic input resolution. FastVLM models using dynamic resolution employs a simple 2$\times$2 grid, with tile size set to 1024. \textdagger - performance numbers reported from~\cite{tong2024cambrian1}. 
For VLMs that use multiple vision encoders, the size of each encoder is listed independently, for TTFT, the latency from each encoder is summed up.
}\label{tab:main_result}
\vspace*{-5pt}
\end{table*}

\section{Experiments} \label{sec:experiments}

\textbf{Training Setup.}
For all the ablations presented in \cref{sec:architecture}, we follow the 2-stage setup described in LLaVA-1.5~\citep{liu2023improvedllava} with Vicuna-7B~\cite{zheng2023judging} as the LLM decoder, unless mentioned otherwise. During the first stage, only the projector is trained using LLaVA-558K alignment dataset for one epoch, with a batch size of 256 and a learning rate of $10^{-3}$. At this stage, the input image resolution matches the backbone pretraining resolution (e.g., 256 for FastViT and 224 for \oursImgEnc{}). In the second stage, we use LLaVA-665K supervised finetuning dataset, training the models for one epoch and tuning all the modules, i.e., vision encoder, projector and the LLM. At this stage, the input image resolution is set to the target resolution.

In \cref{sec:experiments}, we present results with different LLM decoders, primarily with Qwen2-0.5B/1.5B/7B model family~\cite{qwen2} (chat variant) and Vicuna-7B model~\cite{zheng2023judging}. We report results in two training setups, the first one is the 2-Stage setup introduced in LLaVA-1.5.  
For the second training setup, we follow the current trend in literature~\cite{li2024llavanext-ablations, mckinzie2024mm1methodsanalysis} of training the VLMs in 3 stages, i.e. Stage 1 for training the connector, Stage 1.5 for resolution scaling and Stage 2 for visual instruction tuning. Information on datasets used in these stages can be found in Sec.\hyperlink{supp_datasets_info}{\ref*{sec:datasets_info}}.  
In this setup, the input image resolution is set to the backbone pretraining resolution for Stage 1 and adjusted to the target resolution for the following two stages. In both setups, the vision encoder and LLM are frozen only in stage 1, while all modules are finetuned in the remaining stages. For the best setup, we further finetune the model from Stage 2 with high quality instruction tuning dataset with chain-of-thought reasoning from~\cite{guo2024mammothvlelicitingmultimodalreasoning} and call this Stage 3. Details of this setup is elaborated in Sec.\hyperlink{supp_training_setup}{~\ref*{sec:training_setup}} and Sec.\hyperlink{supp_datasets_info}{~\ref*{sec:datasets_info}}. We publicly release
R4, R12, and R41 checkpoints from Stage 2 training,
as well as the R5, R13, and R42 checkpoints from Stage 3
training, as part of our open-sourced codebase.

All FastVLM models reported in the paper are trained on a single node with 8×
NVIDIA H100-80GB GPUs. Stage 1 training of VLM is quick, taking roughly 30 minutes to train with a Qwen2-7B decoder. Stage 1.5 and Stage 2 training runs are dependent on input resolution. For an input resolution of 1024$\times$1024, Stage 1.5 takes 77 hours and Stage 2 takes 8 hours. The reported wall clock times correspond to the following datasets used in these stages: 15 million samples in Stage 1.5 and 1.1 million samples in Stage 2.

\textbf{Evaluation.}
We evaluate the models on the mainstream benchmarks of GQA~\cite{hudson2019gqa}, ScienceQA~\cite{lu2022learnsqa}, TextVQA~\cite{singh2019TextVQA}, POPE~\cite{li2023POPE}, LLaVA-in-the-wild~\cite{liu2023llava}, VQAv2~\cite{goyal2017vqav2}, MMVet~\cite{yu2023mmvet},  MMMU~\cite{yue2023mmmu}, DocVQA~\cite{mathew2021docvqa} and SeedBench~\cite{li2023seedbench}. For GQA, ScienceQA, TextVQA, POPE and LLaVA-in-the-wild benchmarks, we use the official evaluation from LLaVA~\cite{liu2023llava}. For the remaining evaluations we use  lmms-eval~\cite{zhang2024lmmsevalrealitycheckevaluation} library v0.2.2. We use the default settings for all the evaluations and lmms-eval defaults to 0613 version of GPT for evaluations that rely on GPT as a judge.

For ablations presented in \cref{sec:architecture}, we report GQA, TextVQA, POPE, DocVQA and SeedBench. GQA and SeedBench are general knowledge benchmarks, DocVQA and TextVQA represent text-rich evaluations and POPE is a hallucination benchmark. Together these benchmarks provide diversity and are quick to evaluate for ablations. Most importantly, they exhibit lower variance to different initializations and under probabilistic decoding setting. We report the variance for all the evals for different initialization in 
Sec. \hyperlink{supp_eval_variance}{\ref*{sec:eval_variance}}. The standard deviation across the 5 selected metrics is less than 0.5. We call the average of these 5 benchmarks Avg-5, and use it as a reliable signal for our analysis. The empirical standard deviation estimate for Avg-5 is 0.1.

\textbf{Benchmarking.}
We benchmark all the models on a MacBook Pro with the M1 Max chip and 32GB RAM. The image encoder is converted to a Core ML package file using coremltools v7.2 and benchmarked on the neural engine using XCode 15.4 (15F31d). The LLM is benchmarked on the MacBook Pro GPU using MLX~\cite{mlx2023}. The model is first converted using \texttt{mlx\_lm.convert} tool, which converts the models on huggingface to the MLX format and casts the tensors to FP16. The prefilling latency is estimated using \texttt{mlx\_lm.cache\_prompt} tool~\cite{mlx2023}. Time-To-First-Token (TTFT) is estimated by adding the image encoding latency at a specific resolution to the LLM prefilling latency for the associated visual tokens. 

\subsection{Comparison with state-of-the-art}
In \cref{tab:main_result}, we compare FastVLM with recently published methods. The training setup can vary widely between works. For each, we report the LLM decoder and the sizes of the instruction tuning and pretraining datasets used to train the respective VLMs, to facilitate a fair comparison. 

\textbf{Hierarchical Backbones.}
When we compare FastVLM (R20) with ConvLLaVA~\cite{ge2024convllava} (R18), with the same LLM and similar training data size, our model obtains +8.4\% better performance on TextVQA and +12.5\% better performance on DocVQA while being 22\% faster. The gap widens at higher resolution, where FastVLM (R28 and R29) achieves superior performance on wide range of benchmarks while being 2$\times$ faster than ConvLLaVA (R26), with the same LLM decoder.

\textbf{Dataset Scaling.}
When scaling the pretraining dataset by incorporating an intermediate pretraining stage for resolution scaling with 15M samples, FastVLM (R21) matches or surpasses MM1~\citep{mckinzie2024mm1methodsanalysis} (R38) across a wide range of benchmarks. Remarkably, FastVLM achieves this performance while generating 5$\times$ fewer visual tokens. With an input resolution of 1024$\times$1024 and a larger instruction tuning dataset of size 12.5M, FastVLM (R41) outperforms MM1 (R38) and LLaVA-NeXT (R39) across various benchmarks, including text-rich evaluations, like TextVQA and DocVQA, which are sensitive to input resolution and number of visual tokens.
The gap widens when further scale up input resolution using AnyRes, more details in Sec. \hyperlink{supp_anyres_results}{\ref*{sec:anyres_results}}. We provide details of the dataset splits in Sec.\hyperlink{supp_datasets_info}{~\ref*{sec:datasets_info}}.

\textbf{Multiple Vision Encoders.} Recently, MiniGemini~\cite{li2024mgm} and Cambrian-1~\cite{tong2024cambrian1} introduced models that rely on multiple vision encoders. In \cref{tab:main_result}, we compare FastVLM (R40), which uses a single vision encoder with methods that use multiple encoders and trained on similarly scaled visual instruction tuning dataset. In Cambrian-1~\cite{tong2024cambrian1} (R44), vision encoding contributes 3.2$\times$ more than LLM prefilling to the total time-to-first-token of approximately 5 seconds (detailed breakdown is provided in \cref{tab:benchmark_results}). FastVLM (R40) outperforms Cambrian-1 (R44) when trained on a similar visual instruction tuning dataset, while being 7.9$\times$ faster. By scaling the instruction tuning dataset to 12.5M, FastVLM (R41) achieves superior performance over Cambrian-1 (R44) with 2.3$\times$ fewer visual tokens, even on text-rich evaluations (see \cref{tab:text_rich_benchmarks}) that are sensitive to the number of visual tokens.

\textbf{Effect of Decoder.}
VLM performance also depends on the quality of LLM, as demonstrated in prior studies, like~\cite{li2024llavanext-strong}. By switching from Vicuna-7B (R21, R29) to Qwen2~\cite{qwen2,qwen2.5} models (R22, R30), we see improvement in performance across all the benchmarks. The improvements are significant on MMVet, LLaVA-in-the-wild and MMMU benchmarks. With Qwen2-0.5B as the LLM decoder, FastVLM (R4) outperforms LLaVA-OneVision~\cite{li2024llava} (R2) while being 85$\times$ faster. This result underscores the quality of our vision encoder, as both models use the same LLM decoder, while \oursImgEnc{} is 3.4$\times$ smaller compared to SigLIP-SO400M~\cite{zhai2023siglip}.

\begin{table*}[t]
\centering
\resizebox{2.08\columnwidth}{!}{
    \setlength{\tabcolsep}{2pt}
    \begin{tabular}{l|ccccc|ccccccccc}
    \toprule
    Row & \multirow{2}{*}{Method} & Vision & \multirow{2}{*}{LLM} & \#Visual & Vis. Enc. & \multirow{2}{*}{ChartQA}  & \multirow{2}{*}{OCRBench} & \multirow{2}{*}{TextVQA} & \multirow{2}{*}{DocVQA} & \multirow{2}{*}{InfoVQA} & \multirow{2}{*}{MMMU} & \multirow{2}{*}{AI2D} & \multirow{2}{*}{SQA} & Seed  \\
    Ann. & & Encoder & &  Tokens & Size(M)$\downarrow$  & &  &  &  &  & & & & Bench$^I$\\

    \midrule
    \multicolumn{14}{c}{0.5B Model Comparison} \\
    \midrule

    R1 & SmolVLM2-0.5B~\cite{marafioti2025smolvlm}\textsuperscript{\textdagger} & ViT-B & SmolLM2 & 1088$^*$ & \textbf{93} & 62.8 & \textbf{61.0} & 60.2 & 70.5 & 25.5 & \textbf{33.7} & 59.2 & 80.0 & 62.2 \\

    \rowcolor{lightmintbg} R2 & \textbf{FastVLM-0.6B (Ours)} & \oursImgEnc{} & Qw.2 & \textbf{256} & 125 & \textbf{71.4} & 55.8 & \textbf{65.8} & \textbf{79.1} & \textbf{43.3} & 33.6 & \textbf{66.0} & \textbf{82.0} & \textbf{69.3}  \\

    \midrule
    \multicolumn{14}{c}{1-3B Model Comparison} \\
    \midrule

    R3 & SmolVLM2-2.2B~\cite{marafioti2025smolvlm}\textsuperscript{\textdagger} & ViT-SO400M & SmolLM2 & 2106$^*$ & 430 & 68.7 & \textbf{72.9} & \textbf{73.0} & 80.0 & 37.8 & \textbf{42.0} & 70.0 & 89.6 & 71.3 \\

    R4 & MolmoE A1B-7B~\cite{molmo2024}\textsuperscript{\textdagger} & ViT-SO400M   & OLMoE.  & 1728 & 430 & 48.0 & 54.7 & 61.5 & 77.7 & - & 33.9 & 71.0 & 87.5 & 67.9 \\

    R5 & MM1.5-1B~\cite{zhang2024mm15methodsanalysis} & ViT-H   &   -  & 1440  &  632 & 67.2 & 60.5 & 72.5 & 81.0 & 50.5 & 35.8 & 59.3 & 82.1 & 70.2 \\

    R6 & FlorenceVL-3B~\cite{chen2024florence} & DaViT & Phi-3 & 576 & 770 & \textbf{70.7} & 63.0 & 69.1 & 82.1 & \textbf{51.3} & 41.8 & 73.8 & 84.6 & 70.6\\

    \rowcolor{lightmintbg} R7 & \textbf{FastVLM-1.7B (Ours)} & \oursImgEnc{} & Qw.2 & \textbf{256} & \textbf{125} & 69.5 & 61.2 & 71.2 & \textbf{84.2} & 49.6 & 38.0 & \textbf{76.6} & \textbf{92.9} & \textbf{72.6} \\
    
    \bottomrule
    \end{tabular}%
}
\caption{
\textbf{Comparison with concurrent works.} Models are ordered in the descending order of the number of visual tokens. The number of visual tokens listed for each model is based on the highest supported input resolution for the respective models. $^*$ -  For the number of visual tokens generated by SmolVLM2, we estimated it based on the model configs as there is no explicit mention of this information in the paper. The reported visual tokens correspond to the highest resolution supported by the respective model. \textsuperscript{\textdagger} - models are evaluated using VLMEvalKit~\cite{duan2024vlmevalkit}. The rest of the models are evaluated using lmms-eval~\cite{zhang2024lmmsevalrealitycheckevaluation} library. The FastVLM model used in this table is R5 and R13 from \cref{tab:main_result}  
}\label{tab:concurrent_works}
\vspace*{-5pt}
\end{table*}

\textbf{Concurrent works.}
In \cref{tab:concurrent_works}, we compare FastVLM with concurrent works like SmolVLM2~\cite{marafioti2025smolvlm} and FlorenceVL~\cite{chen2024florence}. FastVLM (R7) performs competitively against SmolVLM2 (R3), particularly on text-rich evaluations such as ChartQA and TextVQA, while surpassing it on the DocVQA and InfoVQA benchmarks. FastVLM (R7) accomplishes this using 8.2$\times$ fewer visual tokens than SmolVLM2 (R3). Additionally, FastVLM (R7) outperforms FlorenceVL (R6) on benchmarks such as TextVQA and DocVQA, while using 2.3$\times$ fewer visual tokens and a 6.2$\times$ smaller vision encoder. On knowledge based evaluations like AI2D and ScienceQA, FastVLM (R7) outperforms all other competing models. At the smallest scale, FastVLM (R2) outperforms a similar sized SmolVLM2 model (R1) on a wide range of benchmarks while using 4.3$\times$ fewer visual tokens.

%% file: sec/5_conclusion.tex
\section{Conclusion}
In this work, we introduced FastVLM, which leverages the \oursImgEnc{} vision backbone for efficient encoding of high-resolution inputs. \oursImgEnc{} has a hybrid architecture, is pretrained on reinforced image-text data, and outputs a substantially reduced number of visual tokens with minimal accuracy sacrifice. FastVLM has competitive performance with prior works across a wide range of VLM benchmarks while improving efficiency in both time-to-first-token and the number of parameters in the vision backbone. Rigorous benchmarking on an M1 MacBook Pro demonstrates that FastVLM achieves a state-of-the-art resolution-latency-accuracy trade-off compared to existing works.

%% file: sec/6_acknowledgement.tex
\section*{Acknowledgement}
We thank David Koski, Christopher Webb, Matt Biddulph, Nick Henderson, Megan Maher Welsh, Jamie Cheng, and Jerremy Holland for their valuable contributions to prototyping the demo app. We are also grateful to Angelos Katharopoulos and Awni Hannun for their insightful feedback and suggestions on benchmarking. Additionally, we thank Jen-Hao Rick Chang and Vishwanath Sindagi for their helpful feedback and recommendations.

%% file: sec/X_suppl.tex
\clearpage
\setcounter{page}{1}
\maketitlesupplementary

\renewcommand*{\thesection}{\Alph{section}}
\setcounter{section}{0}

\input{appendix/hyperparams}
\input{appendix/additional_results}
\input{appendix/datasets}

\input{appendix/visualizations}


%% file: appendix/hyperparams.tex
\section{Training Setup}\hypertarget{supp_training_setup}{}\label{sec:training_setup}

For experiments presented in \cref{tab:fastvit_as_enc}, \cref{tab:multiscale_features}, \cref{tab:backbone_ablation},  \cref{tab:comp_token_pruning}, we perform 2-stage training with the hyperparameters listed in \cref{tab:hyp-llava}. The model is trained for a single epoch in all the stages. Note, in \cref{tab:comp_token_pruning}, we do not re-train other token pruning works, we simply report the performance of the respective methods as they adhere to the 2-stage training setup described in \cref{tab:hyp-llava}, which was originally introduced in LLaVA-1.5~\cite{liu2023improvedllava}.

To showcase our model's performance in the presence of additional dataset, we scale both pretraining and instruction tuning datasets in \cref{sec:experiments}. For results presented in R15, R19, R20, R27, R28 in \cref{tab:main_result}, we still perform 2-stage training described in \cref{tab:hyp-llava}, for R20 and R28, we use instruction tuning dataset of size 1.1 million samples in Stage-2. For results presented in R3, R4, R7, R8, R11, R12, R16, R21, R22, R23, R29, R30, R40 and R41 we scale-up both instruction tuning dataset and pretraining dataset. We also introduce and additional stage of pretraining with the scaled-up dataset as described in \cref{tab:hyp-scaled}. Details of 1.1 million, 6.5 million and 12.5 million instruction tuning dataset is presented in Sec.\hyperlink{supp_datasets_info}{\ref*{sec:datasets_info}}. For results presented in R5, R13 and R42, we introduce additional finetuning on high quality instruction tuning dataset from~\cite{guo2024mammothvlelicitingmultimodalreasoning} in Stage 3. We publicly release R4, R12, and R41 checkpoints from Stage 2 training, as well as the R5, R13, and R42 checkpoints from Stage 3 training, as part of our open-sourced codebase.

\begin{table}[ht]
\centering
\resizebox{0.78\columnwidth}{!}{%
\begin{tabular}{l|cc}
\toprule
           & Stage-1              & Stage-2              \\ 
\midrule
Data            & LLaVA-1.5 558K & LLaVA-1.5 665k \\
\midrule
Learning Rate              & 1e-3           & 2e-5           \\
Batch size      & 256            & 128            \\
LR. schedule     & cosine decay   & cosine decay   \\
LR. warmup ratio & 0.03           & 0.03           \\
Optimizer       & AdamW          & AdamW      \\ 
\midrule
Trainable & \multirow{2}{*}{Projector} & Full \\
modules & & Model \\
\bottomrule   
\end{tabular}
}
\caption{2-Stage training setup used in ablations for \cref{sec:architecture}.}
\label{tab:hyp-llava}
\end{table}

\begin{table}[ht]
\resizebox{0.99\columnwidth}{!}{%
\setlength{\tabcolsep}{4pt}
\begin{tabular}{l|cccc}
\toprule
                 & Stage-1        & Stage-1.5       & Stage-2  & Stage 3     \\ 
\midrule
\multirow{2}{*}{Data}  & \multirow{2}{*}{LLaVA-1.5 558K} & Recap-CC3M +    & 1.1M / 6.5M &  \multirow{2}{*}{10.6M} \\
                &                & Recap-CC12M~\cite{li2024llavanext-ablations} & 12.5M & \\
\midrule
Learning Rate              & 1e-3           & 2e-5         & 2e-5 & 2e-5      \\
Batch size      & 256            & 128          & 128      & 128 \\
LR. schedule     & cosine decay   & cosine decay & cosine decay & cosine decay     \\
LR. warmup ratio & 0.03           & 0.03         & 0.03    & 0.03  \\
Optimizer       & AdamW          & AdamW        & AdamW  & AdamW\\ 
\midrule
Trainable & \multirow{2}{*}{Projector} & Full & Full & Full \\
modules & & Model & Model & Model\\
\bottomrule   
\end{tabular}
}
\caption{4-Stage training setup. We report performance after Stage-2 and Stage-3 of training in \cref{tab:main_result}. Please note, both Stage-2 and Stage-3 are visual instruction tuning stages.}
\label{tab:hyp-scaled}
\end{table}

\section{Architecture Details}\hypertarget{supp_arc_details}{}\label{sec:arc_ablation}

The patch embedding layers shown in \cref{fig:arch}, consists of 7$\times$7 depthwise convolutions with~\cite{MobileOne} style train-time overparameterization, followed by 1$\times$1 pointwise convolution. The stride for 7$\times$7 depthwise convolution is set to 2 in order to downsample the input tensor. In~\cite{mobileclip2024}, squeeze-excite layers were incorporated into this block; however, we found them to negatively impact inference latency, especially for high image resolutions, so we opted not to include them in our model. We use the same ConvFFN layer defined in~\cite{vasu2023fastvit}, i.e. 7$\times$7 depthwise convolutions preceding a typical FFN layer. The stem downsamples the input tensor by factor of 4 on each side, and each patch embedding layer downsamples the input tensor by a factor 2. Although recent architectures like ViTamin~\cite{chen2024vitamin} recommend an overall downsampling factor of only 16, \oursImgEnc{} incorporates an additional patch embedding layer compared to FastViT, resulting in an overall downsampling factor of 64× for the input tensor. In each stage, we increase the number of channels by a factor of 2 as done in FastViT and other convolutional and hybrid transformer architectures. This results in a Stage-5 with the widest MLP layers in the architecture, performing self-attention on an input tensor which is downsampled by a factor of 64.

\subsection{Naive Scaling}
In order to scale the model size of FastViT, we simply increased the embedding dimensions per stage to \texttt{[128, 256, 512, 1024]}, and set the number of layers per stage to \texttt{[2, 12, 16, 6]}. Patch embedding layers in each stage use squeeze-excite layers and the MLP expansion ratio is set to 3.0, following the design in~\cite{mobileclip2024}. 

%% file: appendix/additional_results.tex
\begin{table*}[t]
\centering
\resizebox{1.5\columnwidth}{!}{
    \begin{tabular}{l|ccc|cc|ccc}
    \toprule
    Row & \multirow{2}{*}{Method} & Vision & \multirow{2}{*}{LLM} & Input & \#Visual & Vis. Enc. & Vision Enc. & LLM \\
    Ann. & & Encoder & & Res. &  Tokens & Size(M)$\downarrow$  & Latency(ms)$\downarrow$ & Prefilling(ms)$\downarrow$ \\
    \midrule
    \multicolumn{9}{c}{0.5B Model Comparison} \\
    \midrule

    R1 & nanoLLaVA & ViT-SO400M & Qw.1.5 & 384 & 729 & 430 & 272.1 & 263.3 \\

    R2 & LLaVAOV~\cite{li2024llava}$^*$ & ViT-SO400M & Qw.2 & 1152 & 7290 & 430 & 2721.4 & 11402.4 \\ 

    \rowcolor{lightmintbg} R3 & \textbf{FastVLM (Ours)} & \oursImgEnc{} & Qw.2 & 1024 & 256 & \textbf{125} & \textbf{116.3} & 50.5 \\

    \rowcolor{lightmintbg} R3 & \textbf{FastVLM (Ours)$^*$} & \oursImgEnc{} & Qw.2 & 2048 & 1280 & \textbf{125} & 581.5 & 336.4 \\

    \midrule
    \multicolumn{9}{c}{1-2B Model Comparison} \\
    \midrule

    R4 & MobileVLMv2~\cite{chu2024mobilevlm}& ViT-L/14 & ML. & 336 & 144 & 304 & 127.4 & 458 \\

    \rowcolor{lightmintbg} R5 & \textbf{FastVLM (Ours)} & \oursImgEnc{} & Qw.2 & 768 & 144 & \textbf{125} & \textbf{54.8} & 97.1 \\

    \midrule

    R6 & DeepSeekVL~\cite{lu2024deepseekvl} & ViT-SO400M & DS. & 384 & 576 & 430 & 272.1 & - \\

    R7 & MM1~\cite{mckinzie2024mm1methodsanalysis}$^*$ & ViT-H & - & 1344 & 720 & 632 & - & - \\

    \rowcolor{lightmintbg} R8 & \textbf{FastVLM (Ours)} & \oursImgEnc{} & Qw.2 & 1024 & 256 & \textbf{125} & \textbf{116.3} & 116.1 \\

    \rowcolor{lightmintbg} R8 & \textbf{FastVLM (Ours)$^*$} & \oursImgEnc{} & Qw.2 & 2048 & 1280 & \textbf{125} & 581.5 & 681.7 \\

    \midrule
    \multicolumn{9}{c}{7B Model Comparison} \\
    \midrule

    R9 & InstructBLIP~\cite{instructblip} & ViT-g/14 & Vic. & 224 & 32 & 1012 & 149.5 & 152.1 \\ 
    
    \rowcolor{lightmintbg} R11 & \textbf{FastVLM (Ours)} & \oursImgEnc{} & Vic. & 256 & 16 & \textbf{125} & \textbf{6.8} & 143.4 \\

    \midrule

    R12 & MobileVLMv2~\cite{chu2024mobilevlm}& ViT-L/14 & Vic. & 336 & 144 & 304 & 127.4 & 332.1 \\

    R13 & ConvLLaVA~\cite{ge2024convllava} & ConvNeXT-L & Vic. & 768 & 144 & 200 & 164.3 & 332.1 \\

    \rowcolor{lightmintbg} R14 & \textbf{FastVLM (Ours)} & \oursImgEnc{} & Vic. & 768 & 144 & \textbf{125} & \textbf{54.8} & 332.1 \\

    \rowcolor{lightmintbg} R17 & \textbf{FastVLM (Ours)} & \oursImgEnc{} & Qw.2 & 768 & 144 & \textbf{125} & \textbf{54.8} & 391.2 \\

    \midrule

    R20 & ConvLLaVA~\cite{ge2024convllava} & ConvNeXT-L & Vic. & 1024 & 256 & 200 & 696.1 & 461.1 \\

    R26 & LLaVA-1.5~\cite{liu2023improvedllava} & \multirow{3}{*}{ViT-L/14} &  \multirow{3}{*}{Vic.} & 336 & 576 & 304 & 127.4 & 1170.0 \\

    R27 & MobileVLMv2~\cite{chu2024mobilevlm}&   &  & 336 & 576 & 304 & 127.4 & 1170.0 \\

    R28 & ShareGPT4V~\cite{chen2023sharegpt4v} &   &  & 336 & 576 & 304 & 127.4 & 1170.0 \\
    
    R29 & ViTamin~\cite{chen2024vitamin} & ViTamin-L & Vic. & 384 & 576 & 333 & 137.6 & 1170.0 \\

    R30 & ConvLLaVA~\cite{ge2024convllava} & ConvNeXT-L & Vic. & 1536 & 576 & 200 & 1569.7 & 1170.0 \\

    R31 & VILA~\cite{lin2023vila} & ViT-L/14 & L-2  & 336 & 576 & 304 & 127.4 & 1169.5 \\
    
    R33 & MM1~\cite{mckinzie2024mm1methodsanalysis}$^*$ & ViT-H & -  & 1344 & 720 & 632 & - & - \\

    R34 & LLaVA-NeXT$^*$ & ViT-L/14 & L-3 & 672 & 2880 & 304 & 637.0 & 19709.7 \\

    \rowcolor{lightmintbg} R21 & \textbf{FastVLM (Ours)} & \oursImgEnc{} & Vic. & 1024 & 256 & \textbf{125} & \textbf{116.3} & 461.1 \\

    \rowcolor{lightmintbg} R36 & \textbf{FastVLM (Ours)} & \oursImgEnc{} & Qw.2 & 1024 & 256 & \textbf{125} & \textbf{116.3} & 524.5 \\

    \rowcolor{lightmintbg} R36 & \textbf{FastVLM (Ours)$^*$} & \oursImgEnc{} & Qw.2 & 2048 & 1280 & \textbf{125} & 581.5 & 3139.5 \\

    \midrule
    \multicolumn{9}{c}{VLMs with Multiple Vision Encoders and 8B LLM} \\
    \midrule

     \rowcolor{lightpumpkin} & & ConvNeXT-L &  & 1536 &   & 200 & 1569.7 &  \\
     \rowcolor{lightpumpkin} \multirow{-2}{*}{R35} & \multirow{-2}{*}{MiniGemini-HD} & ViT-L/14 & \multirow{-2}{*}{L-3} & 672 & \multirow{-2}{*}{2880} & 304 & 552.6 & \multirow{-2}{*}{19709.7} \\

     \rowcolor{lightpumpkin} & & ViT-SO400M &   & 384  &   & 430 & 272.1 & \\
     \rowcolor{lightpumpkin} & & ConvNeXt-XXL &   & 1024  &   & 846 & 2290.4 &  \\
     \rowcolor{lightpumpkin} & & DINOv2-ViT-L/14 &   & 518 &   & 304 & 1171.5 & \\
     \rowcolor{lightpumpkin} \multirow{-4}{*}{R36} & \multirow{-4}{*}{Cambrian-1~\cite{tong2024cambrian1}} & ViT-L/14 & \multirow{-4}{*}{L-3} & 336 & \multirow{-4}{*}{576} & 304 & 127.4 & \multirow{-4}{*}{1223.6}  \\

    \bottomrule
    \end{tabular}%
}
\caption{
\textbf{Breakdown of prefilling latencies for recent methods.} The models are grouped based on total number of visual tokens. For models that were difficult to export or unavailable, we mark them as '-' in the table. ``Vic.'' refers to Vicuna~\cite{zheng2023judging}, ``Qw.2'' refers to Qwen2~\cite{qwen2} and ``Qw.'' refers to Qwen~\cite{qwen}. ``L-2'' refers to LLaMA-2. ``L-3'' refers to LLaMA-3. ``ML.'' refers to MobileLLaMA~\cite{chu2023mobilevlm, chu2024mobilevlm}. ``DS.'' refers to DeepSeek LLM~\cite{deepseek-llm}. $^*$  For input resolution and visual tokens, we report the highest supported resolution by the respective models as some models like LLaVA-OneVision~\cite{li2024llava} and MM1~\cite{mckinzie2024mm1methodsanalysis} use dynamic input resolution. FastVLM models using dynamic resolution employs a simple 2$\times$2 grid, with tile size set to 1024. For VLMs that use multiple vision encoders, the size of each encoder is listed independently, for TTFT, the latency from each encoder is summed up.
}\label{tab:benchmark_results}
\vspace*{-5pt}
\end{table*}

\section{Additional Results}\label{sec:additional_results}
We present the performance of FastVLM on text-rich benchmarks under various training settings in \cref{tab:text_rich_benchmarks}. FastVLM surpasses MM1 and Cambrian-1 across a wide range of benchmarks by scaling up pretraining and instruction tuning datasets. This result highlights the quality of visual tokens produced by~\oursImgEnc{}, as FastVLM is able to achieve these improvements with 2.8$\times$ less visual tokens than MM1 and with a vision encoder that is 5.1$\times$ smaller.

\begin{table*}[h]
\centering
\resizebox{1.88\columnwidth}{!}{
    \setlength{\tabcolsep}{2pt}
    \begin{tabular}{l|cccc|cc|cc|ccccc}
    \toprule
    Row & \multirow{2}{*}{Method} & Vision & \multirow{2}{*}{LLM} & Data (M) & Input & \#Visual & Vis. Enc. & TTFT & \multirow{2}{*}{ChartQA}  & \multirow{2}{*}{OCRBench} & \multirow{2}{*}{TextVQA} & \multirow{2}{*}{DocVQA} & \multirow{2}{*}{InfoVQA}  \\
    Ann. & & Encoder & & (PT+IT) & Res. &  Tokens & Size(M)$\downarrow$  & (ms)$\downarrow$  &  &   &  & \\
    \midrule
    \multicolumn{14}{c}{0.5B Model Comparison} \\
    \midrule

    R1 & LLaVAOV~\cite{li2024llava}$^*$ & ViT-SO400M & Qw.2 & 4.5+3.2 & 1152 & 7290 & 430 & 14124 & 61.4 & - & - & 70.0 & 46.3 \\ 

    \rowcolor{lightmintbg} R2 & \textbf{FastVLM (Ours)} & \oursImgEnc{} & Qw.2 & 15+12.5 & 1024 & 256 & \textbf{125} & \textbf{166} & 63.4 & 54.9 & 62.9 & 70.4 & 35.8 \\

    \rowcolor{lightmintbg} R3 & \textbf{FastVLM (Ours)} & \oursImgEnc{} & Qw.2 & 15+23.1 & 1024 & 256 & \textbf{125} & \textbf{166} & \textbf{71.4} & 55.8 & \textbf{65.8} & 79.1 & 43.3 \\

    \rowcolor{lightmintbg} R4 & \textbf{FastVLM (Ours)$^*$} & \oursImgEnc{} & Qw.2 & 15+12.5 & 2048 & 1280 & \textbf{125} & 918 & 68.8 & \textbf{59.0} & 65.4 & 82.1 & 49.3 \\

    \rowcolor{lightmintbg} R5 & \textbf{FastVLM (Ours)$^*$} & \oursImgEnc{} & Qw.2 & 15+23.1 & 2048 & 1280 & \textbf{125} & 918 &  64.2 & 55.7 & 65.5 & \textbf{85.3} & \textbf{53.9} \\

    \midrule
    \multicolumn{14}{c}{1-2B Model Comparison} \\
    \midrule

    R6 & MM1~\cite{mckinzie2024mm1methodsanalysis}$^*$ & ViT-H & - & 3000+1.5 & 1344 & 720 & 632 & - & 61.8 & 56.6 & 68.2 & 68.4 & 38.5 \\

    \rowcolor{lightmintbg} R7 & \textbf{FastVLM (Ours)} & \oursImgEnc{} & Qw.2 & 15+12.5 & 1024 & 256 & \textbf{125} & \textbf{233} & 69.6 & 62.9 & 69.0 & 75.6 & 41.7\\

    \rowcolor{lightmintbg} R8 & \textbf{FastVLM (Ours)} & \oursImgEnc{} & Qw.2 & 15+23.1 & 1024 & 256 & \textbf{125} & \textbf{233} & 69.5 & 61.2 & 71.2 & 84.2 & 49.6 \\

    \rowcolor{lightmintbg} R9 & \textbf{FastVLM (Ours)$^*$} & \oursImgEnc{} & Qw.2 & 15+12.5 & 2048 & 1280 & \textbf{125} & 1263 & \textbf{76.4} & \textbf{63.2} & 71.5 & 87.6 & 60.0 \\

    \rowcolor{lightmintbg} R10 & \textbf{FastVLM (Ours)$^*$} & \oursImgEnc{} & Qw.2 & 15+12.5 & 2048 & 1280 & \textbf{125} & 1263 & 71.2 & 62.4 & \textbf{72.9} & \textbf{90.2} & \textbf{62.5} \\

    \midrule
    \multicolumn{14}{c}{7B Model Comparison} \\
    \midrule

    R11 & MM1~\cite{mckinzie2024mm1methodsanalysis}$^*$ & ViT-H & - & 3000+1.5 & 1344 & 720 & 632 & - & 72.6 & 62.6 & 72.8 & 76.8 & 45.5 \\

    R12 & LLaVA-NeXT\textsuperscript{\textdagger}$^*$ & ViT-L/14 & L-3 & - & 672 & 2880 & 304 & 20347 & 69.5 & 49.0 &  64.6 &  72.6 & - \\
    
    \midrule 
    
    \multirow{4}{*}{R13} & \multirow{4}{*}{Cambrian-1~\cite{tong2024cambrian1}} & ViT-L/14 & \multirow{4}{*}{L-3} & \multirow{4}{*}{2.5+7} & 336 & \multirow{4}{*}{576} & 304 & \multirow{4}{*}{5085} & \multirow{4}{*}{73.3} & \multirow{4}{*}{62.4} & \multirow{4}{*}{71.7} & \multirow{4}{*}{77.8} & \multirow{4}{*}{-} \\
     & & ViT-SO400M &  &  & 384  &   & 430 &  & & & & & \\
     & & ConvNeXt-XXL &  &  & 1024  &   & 846 &  & & & & & \\
     & & DINOv2-ViT-L/14 &  &  & 518 &   & 304 &  & & & & & \\
     
     \midrule
   
    \rowcolor{lightmintbg} R14 & \textbf{FastVLM (Ours)} & \oursImgEnc{} & Vic. & 0.5+0.6 & 768 & 144 & \textbf{125} & \textbf{387} & 17.1 & 30.0 & 62.9 & 32.9 & 28.7 \\
    
    \rowcolor{lightmintbg} R15 & \textbf{FastVLM (Ours)} & \oursImgEnc{} & Vic. & 0.5+1.1 & 768 & 144 & \textbf{125} & \textbf{387} & 59.1 & 38.4 & 67.5 & 57.3 & 29.7 \\

    \rowcolor{lightmintbg} R16 & \textbf{FastVLM (Ours)} & \oursImgEnc{} & Vic. & 15+1.1 & 768 & 144 & \textbf{125} & \textbf{387} & 65.4 & 45.3 & 69.4 & 65.5 & 32.0 \\
    
    \rowcolor{lightmintbg} R17 & \textbf{FastVLM (Ours)} & \oursImgEnc{} & Qw.2 & 15+1.1 & 768 & 144 & \textbf{125} & 446 & 69.3 & 45.9 & 69.5 & 66.9 & 34.3 \\

    \rowcolor{lightmintbg} R18 & \textbf{FastVLM (Ours)} & \oursImgEnc{} & Qw.2 & 15+11.9 & 768 & 144 & \textbf{125} & 446 & 74.2 & 59.0 & 72.8 & 72.0 & 44.3 \\

    \midrule

    \rowcolor{lightmintbg} R19 & \textbf{FastVLM (Ours)} & \oursImgEnc{} & Vic. & 0.5+0.6 & 1024 & 256 & \textbf{125} & \textbf{577} & 19.2 & 29.3 & 64.4 & 35.6 & 28.9 \\
    
    \rowcolor{lightmintbg} R20 &\textbf{FastVLM (Ours)} & \oursImgEnc{} & Vic. & 0.5+1.1 & 1024 & 256 & \textbf{125} & \textbf{577} & 61.0 & 38.3 & 67.4 & 62.8 & 32.0 \\

    \rowcolor{lightmintbg} R21 & \textbf{FastVLM (Ours)} & \oursImgEnc{} & Vic. & 15+1.1 & 1024 & 256 & \textbf{125} & \textbf{577} & 66.9 & 47.1 & 70.6 & 72.4 & 34.7 \\

    \rowcolor{lightmintbg} R22 & \textbf{FastVLM (Ours)} & \oursImgEnc{} & Qw.2 & 15+1.1 & 1024 & 256 & \textbf{125} & 641 & 71.0 & 49.7 & 72.1 & 73.3 & 37.5 \\
   
    \rowcolor{lightmintbg} R23 & \textbf{FastVLM (Ours)} & \oursImgEnc{} & Qw.2 & 15+6.5 & 1024 & 256 & \textbf{125} & 641 & 76.6 & 52.9 & 73.1 & 78.7 & 44.2 \\

    \rowcolor{lightmintbg} R24 & \textbf{FastVLM (Ours)} & \oursImgEnc{} & Qw.2 & 15+11.9 & 1024 & 256 & \textbf{125} & 641 & 77.0 & 63.3 & 74.8 & 78.9 & 49.7 \\

    \rowcolor{lightmintbg} R25 & \textbf{FastVLM (Ours)} & \oursImgEnc{} & Qw.2 & 15+12.5 & 1024 & 256 & \textbf{125} & 641 & 77.5 & 65.7 & 73.4 & 82.7 & 51.2 \\

    \rowcolor{lightmintbg} R26 & \textbf{FastVLM (Ours)} & \oursImgEnc{} & Qw.2 & 15+23.1 & 1024 & 256 & \textbf{125} & 641 & \textbf{78.1} & \textbf{67.3} & \textbf{76.1} & \textbf{88.0} & \textbf{55.1} \\

    \midrule

    \rowcolor{lightmintbg} R27 & \textbf{FastVLM (Ours)$^*$} & \oursImgEnc{} & Qw.2 & 15+12.5 & 2048 & 1280 & \textbf{125} & 3721 & \textbf{82.4} & \textbf{67.3} & 76.6 & 92.3 & 68.3 \\

    \rowcolor{lightmintbg} R28 & \textbf{FastVLM (Ours)$^*$} & \oursImgEnc{} & Qw.2 & 15+23.1 & 2048 & 1280 & \textbf{125} & 3721 & 74.6 & 65.7 & \textbf{77.1} & \textbf{94.5} & \textbf{71.1} \\

    \bottomrule
    \end{tabular}%
}
\caption{
\textbf{Comparison with recent methods on text-rich benchmarks.} The models are grouped based on total number of visual tokens. ``-'' indicates that performance was not reported in the respective paper. For the dataset column, ``-'' indicates that the dataset size for pretraining (``PT'') or instruction tuning (``IT'') is not explicitly mentioned in the respective paper. For methods that have more than 2 stages of training, we report the total samples used for all the pretraining stages as part of ``PT''. ``TTFT'' means time to first token (the sum of the vision encoder latency and the LLM prefilling time), we report latency only for models that are publicly available and in a format favorable to MLX~\citep{mlx2023} ``Vic.'' refers to Vicuna~\cite{zheng2023judging}, ``Qw.2'' refers to Qwen2~\cite{qwen2}. ``L-3'' refers to LLaMA-3. * - For input resolution and visual tokens, we report the highest supported resolution by the respective models that use dynamic input resolution. \textdagger - performance numbers reported from~\cite{tong2024cambrian1}. For VLMs that use multiple vision encoders, the size of each encoder is listed independently, for TTFT, the latency from each encoder is summed up.
}\label{tab:text_rich_benchmarks}
\vspace*{-5pt}
\end{table*}

\subsection{Dynamic Resolution (AnyRes) Results}\hypertarget{supp_anyres_results}{}\label{sec:anyres_results}
From \cref{fig:res_compare}, it is evident that VLMs prefer visual encoding with fewer semantic breaks. Variants with more tiles typically underperform compared to those with fewer tiles and a static resolution. To scale up input resolution, we train variants of FastVLM with support for dynamic input resolution, where we use a tile size of 1024$\times$1024 and use a simple 2$\times$2 grid. This enables the model to process a peak input resolution of 2048$\times$2048 using only 4 tiles, unlike models like InternVL2~\cite{chen2024far} which uses roughly 36 tiles to process images of resolution 2688$\times$2688. We report performance of FastVLM (R4, R5, R9, R10, R27, R28) with dynamic resolution support on text-rich benchmarks in \cref{tab:text_rich_benchmarks}.

\subsection{CVBench and MathVista Results}\hypertarget{supp_cvbench_mathvista}{}\label{sec:cvbench_mathvista}
Results in \cref{tab:cvbench_mathvista} show that, in comparison to Cambrian-1~\cite{tong2024cambrian1}, FastVLM is significantly better on MathVista~\cite{lu2024mathvista} and competitive on CVBench~\cite{tong2024cambrian1}, even though we have a single backbone and significantly fewer tokens. Results on both CVBench and MathVista benchmarks further improve as we scale the SFT dataset by including LLaVA-OneVision~\cite{li2024llava}.

\vspace*{-5pt}
\begin{table}[h]
\centering
\resizebox{0.99\columnwidth}{!}{%
    \setlength{\tabcolsep}{2pt}
    \begin{tabular}{l|ccccc|ccc}
    \toprule
    \multirow{2}{*}{Method}  & LLM & Data (M) & Input & \#Visual & Latency & {CVBench} & {CVBench} & Math \\
     & Decoder & (PT+IT) & Res.  &  Tokens & Enc.(ms)  & 2D & 3D & Vista\\
    \midrule
    Cambrian-1         & LLama3-8B & 2.5+7 &Mult.& 576 & 3861.4 & 72.3  & 72.0  & 49.0 \\
    \textbf{FastVLM}   & Qwen2-7B  & 15+6.5 & 1024  & 256 & 116.3 & 71.8  & 69.6  & 57.6 \\
    \midrule
    \textbf{FastVLM}   & Qwen2-7B & 15+11.9  & 768   & 144 & 54.8 & 75.9  & 79.3  & 64.6 \\
    \textbf{FastVLM}   & Qwen2-7B & 15+11.9 & 1024  & 256 & 116.3 & 76.7  & 80.9  & 64.8 \\
    \bottomrule
    \end{tabular}%
}\vspace*{-5pt}
\caption{\small
\textbf{Evaluation on CVBench and MathVista.} ``6.5M'' instruction tuning dataset is from Cambrian-1. ``11.9'' instruction tuning dataset is concatenation of Cambrian-1 and LLaVA-OneVision datasets.
}\label{tab:cvbench_mathvista}
    \vspace*{-10pt}
\end{table}

%% file: appendix/datasets.tex
\section{Datasets}\hypertarget{supp_datasets_info}{}\label{sec:datasets_info}

\subsection{Pretraining Datasets}

For Stage-1 training, we only use LLaVA-1.5 558K~\cite{liu2023improvedllava} dataset. For Stage-1.5 training, we use densely captioned versions of  CC3M~\cite{CC3M} and CC12M~\cite{CC12M} introduced in~\cite{li2024llavanext-ablations}. The total size of this dataset is 15 million image-text pairs. 
We generated 300 generic questions, such as ``What is in this photo?''. For each (image, dense-caption) pair, we randomly selected a generic question to form a triplet of (question, image, dense-caption). With a 0.5 probability, we placed the image’s special token \texttt{<image>} either before or after the question.
From recent works like~\cite{li2024llavanext-ablations, tong2024cambrian1, mckinzie2024mm1methodsanalysis} and our results in~\cref{tab:main_result}, scaling dataset in Scale-1.5 is beneficial to improve the performance of VLM across a wide range of evaluations. Even though~\oursImgEnc{} is smaller than ViT-L/14 and ViT-H used in ~\cite{li2024llavanext-ablations, mckinzie2024mm1methodsanalysis} respectively, we see similar scaling trends.  

\subsection{Visual Instruction Tuning Datasets}

We use 3 different version of instruction tuning datasets. The smallest scale is LLaVA-1.5 665K dataset~\cite{liu2023improvedllava}. We further scale up this dataset by including training splits of the following datasets; AI2D~\cite{kembhavi2016diagram}, ScienceQA~\cite{lu2022learnsqa}, ChartQA~\cite{masry-etal-2022-chartqa}, COCO~\cite{COCO}, DocVQA~\cite{mathew2021docvqa}, DVQA~\cite{kafle2018dvqa}, GeoQA+~\cite{cao-xiao-2022-augmented}, OCRVQA~\cite{mishraICDAR19}, SegmentAnything~\cite{kirillov2023segany}, SynthDoG-EN~\cite{kim2022donut}, TextVQA~\cite{singh2019TextVQA} and Visual Genome~\cite{Krishna2016VisualGC}. The conversational data for the listed datasets is sourced from~\cite{chen2023internvl}. The total number of samples in this dataset is 1.1 million and is referred to as ``1.1M'' in all the tables. We further scale-up instruction tuning dataset using image-based conversational data from Cambrian-7M~\cite{tong2024cambrian1}, which amounts to 5.4 million samples. Filtered Cambrian-7M~\cite{tong2024cambrian1} is merged with ``1.1M '' dataset to obtain ``6.5M'' instruction tuning dataset. We then append all available single-image instruction tuning data open-sourced by LLaVA-OneVision~\cite{li2024llava} to ``6.5M'' to obtain ``11.9M'' instruction tuning dataset. We then include roughly 0.6M samples from DocMatix~\cite{laurençon2024building} dataset to obtain ``12.5M'' instruction tuning dataset. From \cref{tab:main_result}, we see further improvements in VLM benchmarks when instruction tuning dataset is scaled, following trends exhibited by image encoders much bigger than~\oursImgEnc{}. The best performing models are achieved when Stage 2 models are further fine-tuned on the MammothVL instruction tuning dataset~\cite{guo2024mammothvlelicitingmultimodalreasoning}. Specifically, we filter the corpus to include only single-image instruction tuning data, resulting in ``10.6M'' samples.    

\subsection{Evaluations}\hypertarget{supp_eval_variance}{}\label{sec:eval_variance}

In addition to evaluations listed in \cref{sec:experiments}, we report performance of FastVLM on ChartQA~\cite{masry-etal-2022-chartqa}, OCRBench~\cite{liu2024ocrbenchhiddenmysteryocrbench} and InfoVQA~\cite{mathew2021infographicvqa} to compare FastVLM against recent methods on text-rich benchmarks. In \cref{tab:multi_run}, report performance of FastViT model (with architectural interventions) from multiple training runs and compute the standard deviation of metrics reported in \cref{tab:main_result}. As described in \cref{sec:experiments}, for ablations we are interested in benchmarks that are quick to evaluate and exhibit lower variance to different initializations. From \cref{tab:multi_run}, GQA, TextVQA, POPE, DocVQA and SeedBench fit the criteria. While VQAv2 also exhibits lower variance it is substantially larger and takes long time to evaluate. The standard deviation across the selected metrics is below 0.5, so we use the average of these metrics as a reliable indicator for our analysis in \cref{sec:architecture}.

\begin{table}[t]
\resizebox{0.99\columnwidth}{!}{%
    \setlength{\tabcolsep}{2pt}
    \begin{tabular}{lccccccccc}
    \toprule
    & \multirow{2}{*}{GQA}  & \multirow{2}{*}{SQA}  & \multirow{2}{*}{TextVQA} & \multirow{2}{*}{POPE} & LLaVA & \multirow{2}{*}{MMVet} & \multirow{2}{*}{VQAv2} & \multirow{2}{*}{DocVQA} & Seed\\
    & & & &  &    Bench$^W$ &    &      &        &   Bench$^I$     \\
    \midrule

    & 62.69 & 64.25 & 60.71 & 85.8 & 59.4 & 29.6 & 77.27 & 27.57 & 53.31 \\
    & 62.68 & 64.95 & 60.61 & 86.1 & 60.1 & 31.6 & 77.39 & 28.37 & 53.55 \\
    & 62.69 & 65.64 & 60.68 & 85.3 & 61.4 & 31.1 & 77.31 & 28.26 & 53.46 \\
   \midrule
   Std. & 0.0047
 & 0.57 & 0.041 & 0.33 & 0.83 & 0.85 & 0.049 & 0.35 & 0.099\\
    \bottomrule
    \end{tabular}%
}
\caption{
VLM benchmarks across three independent runs with frozen FastViT image encoder. Training setup is LLaVA-1.5 with Vicuna 7B as LLM. Standard deviation across runs is listed in the bottom row.
}\label{tab:multi_run}
\end{table}

%% file: appendix/visualizations.tex
\section{Qualitative Analysis}\hypertarget{supp_qualitative_analysis}{}\label{sec:qualitative_analysis}
We analyzed failures across benchmarks and found: In text-rich benchmarks (e.g., DocVQA, ChartQA), failures occur when text is too small or precise alignment is needed (e.g., reading tables). In majority of cases where the text is too small, simply using higher input resolution can reduce errors as shown in \cref{tab:res_update}. Some cases require broader general knowledge to obtain a correct response as seen in \cref{tab:llm_update}, in such cases using a bigger LLM reduces failures. Some cases require reasoning about a higher resolution image, in which case using a bigger LLM can decrease failures as seen in \cref{tab:llm_and_res_update}. Some failures result from misjudgment, where correct responses are misclassified by the LLM judge or incorrect labels, we ignore these cases in our analysis.

\begin{table}[h!]
  \begin{minipage}{0.99\linewidth}
\centering
\scalebox{0.88}{
\begin{tabular}{p{2.5cm} p{5.8cm} }
\toprule
\multicolumn{2}{c}{When to Use a Larger LLM} \\
\midrule

\multicolumn{2}{c}{\includegraphics[width=0.7\textwidth]{figures/fa_examples/mmmu_llm_diff_example_1.png}} \\
\multicolumn{2}{c}
{Dataset: MMMU~\cite{yue2023mmmu}} \\

User & What is the common term for the yellow area surrounding the site of an infection? Options: ["I don't know and I don't want to guess", 'Corona', 'Border', 'Halo', 'Toxin zone'] \\
Ground Truth & D \\

\midrule 

FastVLM-0.5B @ 256 & E \red\xmark{} \\

FastVLM-0.5B @ 1024 & E \red\xmark{} \\

FastVLM-1.5B @ 256 & D \green\cmark{} \\

FastVLM-1.5B @ 1024 & D \green\cmark{} \\

\midrule

\multicolumn{2}{c}{\includegraphics[width=0.8\textwidth]{figures/fa_examples/mmmu_llm_diff_example_2.jpeg}} \\
\multicolumn{2}{c}{Dataset: MMMU~\cite{yue2023mmmu}} \\

User & The sinoatrial (SA) node is indicated by \_\_\_. Options: ['A', 'B', 'C', 'D', 'E'] \\
Ground Truth & A \\

\midrule 

FastVLM-0.5B @ 256 & E \red\xmark{} \\

FastVLM-0.5B @ 1024 & D \red\xmark{} \\

FastVLM-1.5B @ 256 & A \green\cmark{} \\

FastVLM-1.5B @ 1024 & A \green\cmark{} \\

\bottomrule
\end{tabular}
}
\captionof{table}{Cases that require broader general knowledge, in which case a larger LLM is preferred.}
\label{tab:llm_update}
  \end{minipage}
\end{table}

\begin{table}[h!]
  \begin{minipage}{0.99\linewidth}
\centering
\scalebox{0.88}{
\begin{tabular}{p{2.5cm} p{5.8cm} }
\toprule
\multicolumn{2}{c}{When to Use Higher Resolution} \\
\midrule

 \multicolumn{2}{c}{\includegraphics[width=0.5\textwidth]{figures/fa_examples/GQA_res_diff_example_2.png}} \\
 \multicolumn{2}{c}{Dataset: GQA~\cite{hudson2019gqa}} \\

User & What is sitting inside the bowls? \\
Ground Truth & squash \\

\midrule 

FastVLM-0.5B @ 256 & Sculpture \red\xmark{} \\

FastVLM-0.5B @ 1024 & Squash \green\cmark{} \\

FastVLM-1.5B @ 256 & Potato \red\xmark{} \\

FastVLM-1.5B @ 1024 & Squash \green\cmark{} \\

\midrule

\multicolumn{2}{c}{\includegraphics[width=0.8\textwidth]{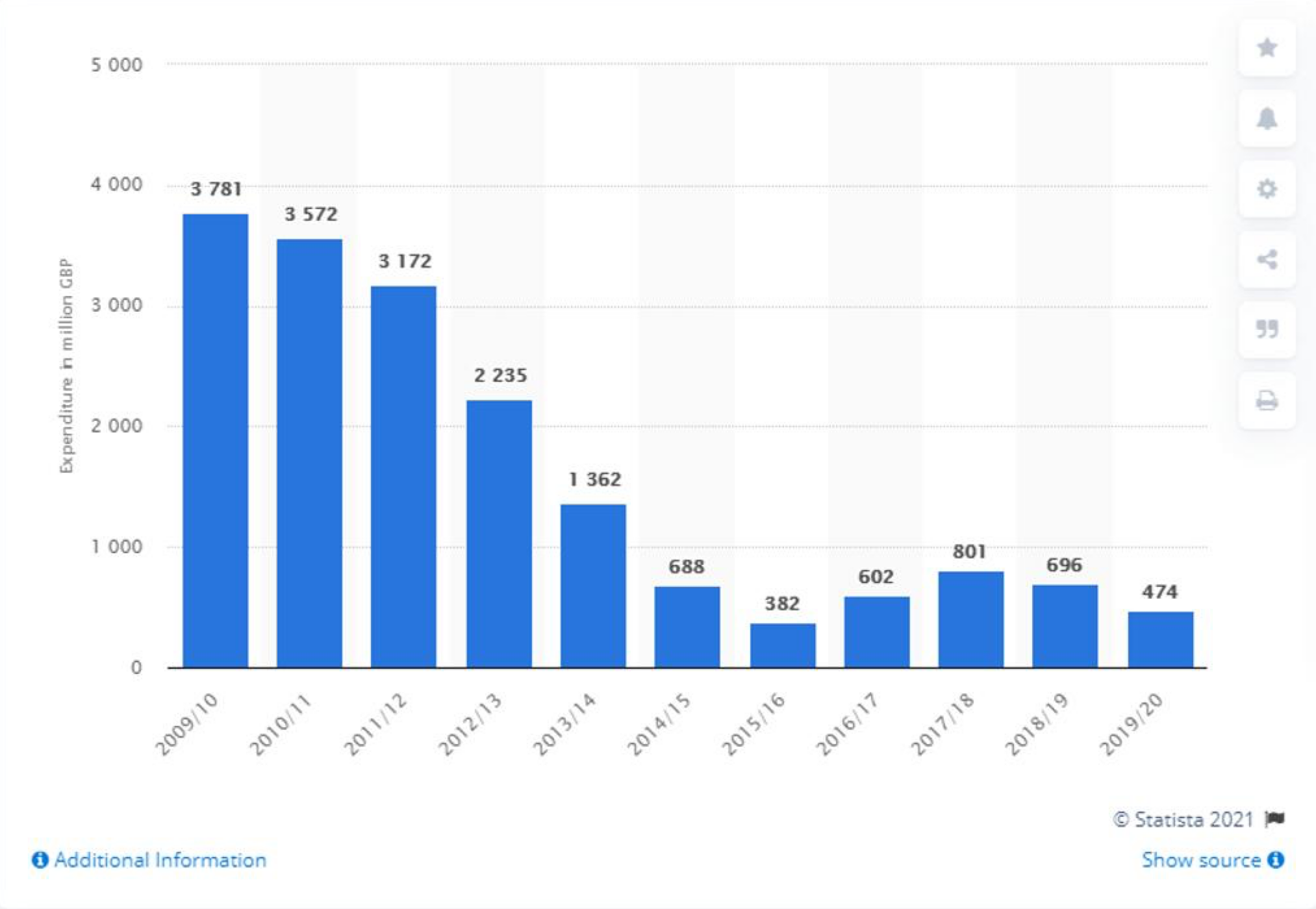}} \\
\multicolumn{2}{c}{Dataset: ChartQA~\cite{masry-etal-2022-chartqa}} \\

User & What was the highest expenditure on foreign military aid in 2009/10? \\
Ground Truth & 3781 \\

\midrule 

FastVLM-0.5B @ 256 & Germany \red\xmark{} \\

FastVLM-0.5B @ 1024 & 3781 \green\cmark{} \\

FastVLM-1.5B @ 256 & 275 \red\xmark{} \\

FastVLM-1.5B @ 1024 & 3781 \green\cmark{} \\

\bottomrule
\end{tabular}
}
\captionof{table}{Cases where increasing the resolution is sufficient to obtain a better response.}
\label{tab:res_update}
  \end{minipage}
\end{table}

\begin{table}[h!]
  \begin{minipage}{0.99\linewidth}
\centering
\scalebox{0.88}{
\begin{tabular}{p{2.5cm} p{5.8cm} }
\toprule
\multicolumn{2}{c}{When to Use Higher Resolution and a Larger LLM} \\
\midrule

 \multicolumn{2}{c}{\includegraphics[width=1.0\textwidth]{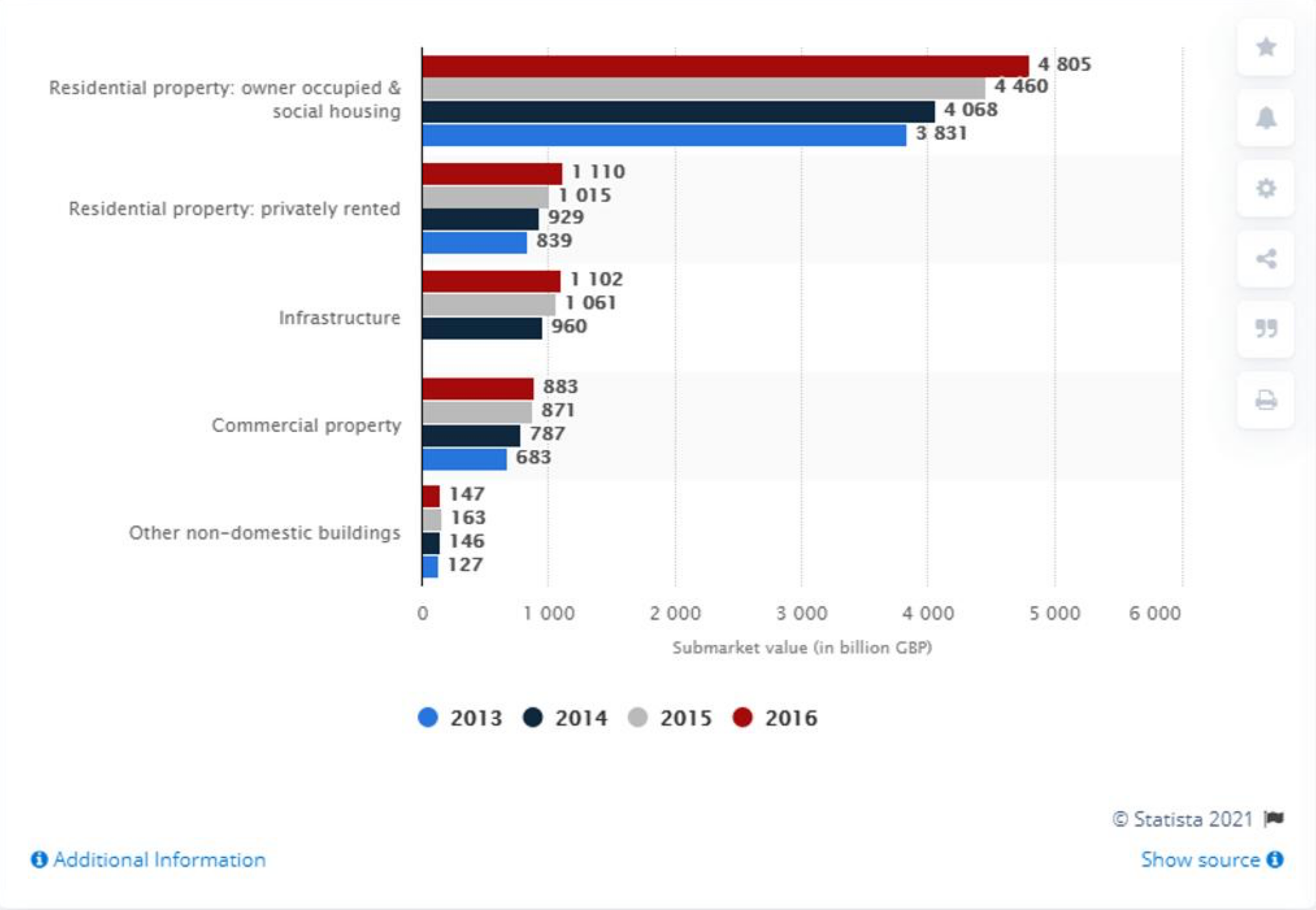}} \\
 \multicolumn{2}{c}{Dataset: ChartQA~\cite{masry-etal-2022-chartqa}} \\

User & What was the value of the commercial property market in 2016? \\
Ground Truth & 883 \\

\midrule 

FastVLM-0.5B @ 256 & 10000 \red\xmark{} \\

FastVLM-0.5B @ 1024 & 871 \red\xmark{} \\

FastVLM-1.5B @ 256 & 1100 \red\xmark{} \\

FastVLM-1.5B @ 1024 & 883 \green\cmark{} \\

\midrule

\multicolumn{2}{c}{\includegraphics[width=0.7\textwidth]{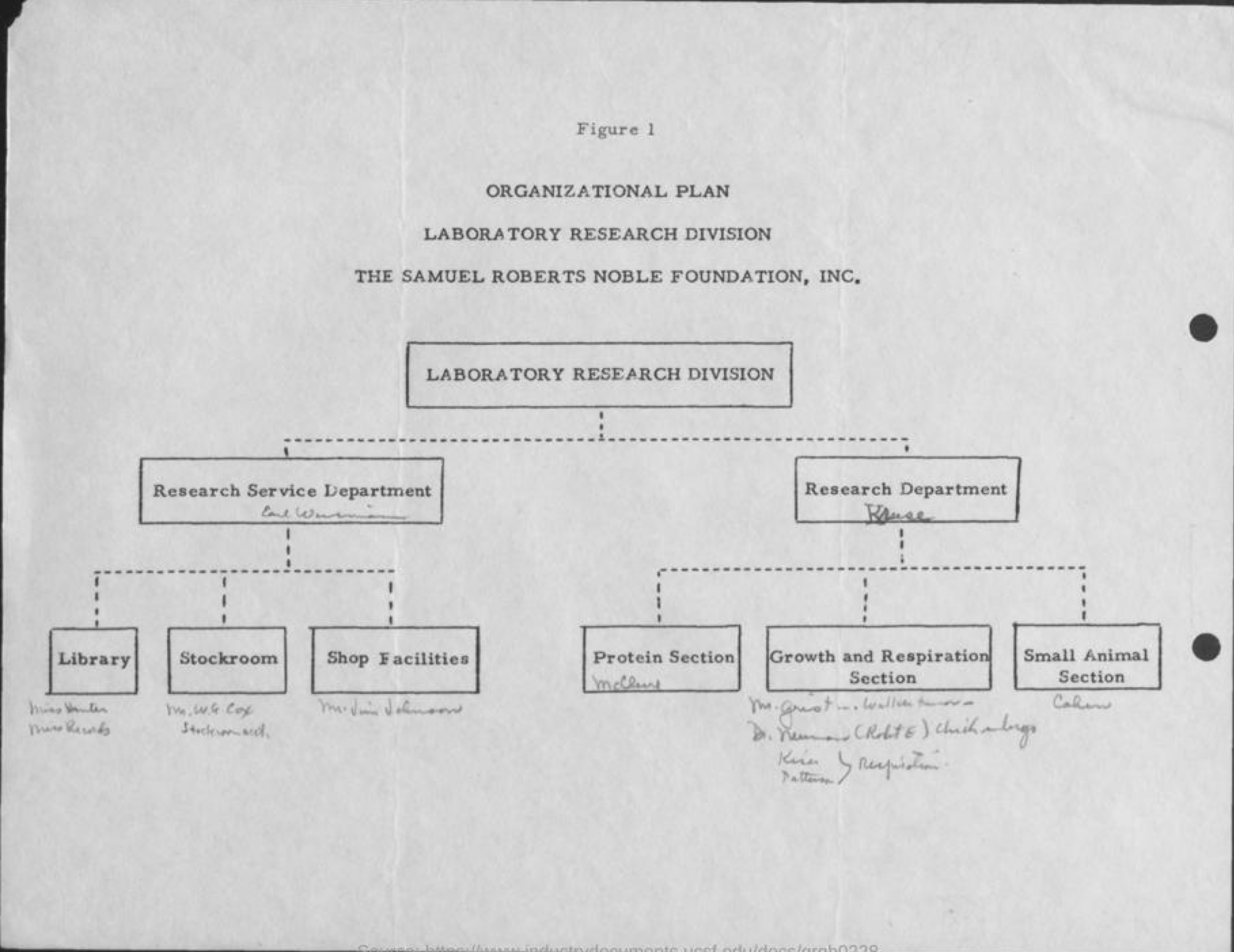}} \\
\multicolumn{2}{c}{Dataset: DocVQA~\cite{mathew2021docvqa}} \\

User & Under which department `Stockroom' is organized? \\
Ground Truth & Research Service Department \\

\midrule 

FastVLM-0.5B @ 256 & Department of Chemistry \red\xmark{} \\

FastVLM-0.5B @ 1024 & Library \red\xmark{} \\

FastVLM-1.5B @ 256 & Research Department \red\xmark{} \\

FastVLM-1.5B @ 1024 & Research Service Department \green\cmark{} \\

\bottomrule
\end{tabular}
}
\captionof{table}{Cases where increasing the resolution alone is not sufficient to obtain a better response. Larger LLM in combination with higher resolution is required to obtain a correct response.}
\label{tab:llm_and_res_update}
  \end{minipage}
\end{table}